%% file: neurips_2025.tex
\documentclass{article}

\usepackage[final]{neurips_2025}

\input{preamble}

\input{notation}
\title{Training-free Online Video Step Grounding}

\author{
Luca Zanella\textsuperscript{1}
\quad
Massimiliano Mancini\textsuperscript{1}
\quad
Yiming Wang\textsuperscript{2}
\quad \vspace{.1em}\\
\textbf{Alessio Tonioni\textsuperscript{3}}
\quad
\textbf{Elisa Ricci\textsuperscript{1,2}} \vspace{1mm} \\
\textsuperscript{1}University of Trento
\quad
\textsuperscript{2}Fondazione Bruno Kessler
\quad
\textsuperscript{3}Google\\[1mm]
{\tt\small \url{https://lucazanella.github.io/baglm/}}
}

\begin{document}

\maketitle

\input{sections/0_abstract}
\input{sections/1_introduction}

\input{sections/2_related}

\input{sections/3_preliminary}

\input{sections/4_method}

\input{sections/5_experiments}

\input{sections/6_conclusion}

\bibliographystyle{plain}
\bibliography{refs} 

\medskip

{
\small

}

\clearpage

\newpage
\appendix
In the Appendix, we first provide the prompts used in \methodshort (Sec.~\ref{sec:app:prompts}). We then present additional experimental analyses (Sec.~\ref{sec:app:ablation}), including the type of prompt used to compute \taskfull scores, the prior used in the update step of our method, and step localization results. We extend these analyses with evaluations on the COIN dataset to demonstrate the generalization of \methodshort across domains and tasks, robustness analyses of LLM-generated dependency matrices on HT-Step, and analyses of step prediction performance by temporal position and by the number of candidate steps. We also describe the computational cost of \methodshort during inference (Sec.~\ref{sec:app:inference}), discuss the broader social impacts of our work (Sec.~\ref{sec:app:broader_impacts}), and list the licenses and URLs for all assets used (Sec.~\ref{sec:app:assets}). Finally, we provide qualitative results (Sec.~\ref{sec:app:qualitatives}).
\input{appendix/a_prompts}

\input{appendix/b_additional_analyses}

\input{appendix/c_inference}
\input{appendix/d_broader_impacts}
\input{appendix/e_assets}
\input{appendix/f_qualitatives}

\end{document}

%% file: preamble.tex
\usepackage[utf8]{inputenc} %
\usepackage[T1]{fontenc}    %
\usepackage{hyperref}       %
\usepackage{url}            %
\usepackage{booktabs}       %
\usepackage{amsfonts}       %
\usepackage{nicefrac}       %
\usepackage{microtype}      %
\usepackage[table]{xcolor}         %
\usepackage{tabularx}

\usepackage{multirow}
\usepackage{graphicx}
\usepackage{amsmath}
\usepackage{pifont}
\usepackage{xspace}
\usepackage{subcaption}
\usepackage{wrapfig}
\usepackage{caption}
\usepackage{colortbl}
\usepackage[most]{tcolorbox}

\makeatletter
\DeclareRobustCommand\onedot{\futurelet\@let@token\@onedot}
\def\@onedot{\ifx\@let@token.\else.\null\fi\xspace}

\def\eg{\emph{e.g}\onedot} 
\def\ie{\emph{i.e}\onedot}

\def\wrt{w.r.t\onedot}

\def\suppmat{Appendix}

\definecolor{DrawioBlue}{RGB}{218,232,252}
\definecolor{DrawioOrange}{RGB}{255,230,204}
\definecolor{DrawioGreen1}{RGB}{204,255,204}
\definecolor{DrawioGreen}{RGB}{213,232,212}
\definecolor{DrawioPurple}{RGB}{225,213,231}
\definecolor{DrawioRed}{RGB}{248,206,204}
\definecolor{DrawioYellow}{RGB}{255, 242, 207}
\definecolor{skyblue}{RGB}{135, 206, 235}
\definecolor{lightcoral}{RGB}{240, 128, 128}

\newcommand{\taskshort}{{VSG}\xspace}
\newcommand{\taskfull}{{Video Step Grounding}\xspace}
\newcommand{\taskfullmin}{{video step grounding}\xspace}

\newcommand{\methodshort}{\textsc{BaGLM}\xspace}
\newcommand{\methodfull}{Bayesian Grounding with Large Multimodal Models\xspace}

\newcommand{\inlineColorbox}[2]{\begingroup\setlength{\fboxsep}{1pt}\colorbox{#1}{\hspace*{2pt}\vphantom{Ay}#2\hspace*{2pt}}\endgroup}

%% file: notation.tex
\newcommand{\textspace}{\mathcal{L}}
\newcommand{\video}{\mathbf{V}}               %
\newcommand{\videospace}{\mathcal{V}}         %
\newcommand{\videosegments}{\mathcal{S}}      %
\newcommand{\segment}{\mathbf{S}}             %

\newcommand{\numsegments}{T}                  %

\newcommand{\numsteps}{K}                     %
\newcommand{\steps}{\mathcal{A}}             %
\newcommand{\step}{a}                         %
\newcommand{\stepvar}{\mathtt{A}}
\newcommand{\Real}{\mathbb{R}}
\newcommand{\sharedspace}{\Real^d}%
\newcommand{\wordssimplex}{\Delta^{|\words|}}
\newcommand{\predsimplex}{\Delta^{\numsteps+1}}
\newcommand{\visualencoder}{f_{\mathtt{vid}}} %
\newcommand{\textencoder}{f_{\mathtt{txt}}}   %
\newcommand{\textdecoder}{f_{\mathtt{dec}}}   %

\newcommand{\beliefn}{\mathtt{bel}}

\newcommand{\dependencymatrix}{\mathbf{D}}
\newcommand{\transitionmatrix}{\mathbf{T}}
\newcommand{\words}{\mathcal{W}}              %

\newcommand{\readiness}{\mathbf{r}}                          %
\newcommand{\validity}{\mathbf{v}}         %
\newcommand{\progress}{\mathbf{progress}}

\newcommand{\hiddenstate}{\mathbf{x}}         %
\newcommand{\Hiddenstate}{\mathbf{X}}         %
\newcommand{\measurement}{\mathbf{z}}         %

%% file: sections/0_abstract.tex
\begin{abstract}
Given a task and a set of steps composing it, \taskfull (\taskshort) aims to detect which steps are performed in a video. Standard approaches for this task require a labeled training set (\eg, with step-level annotations or narrations), which may be costly to collect.
Moreover, they process the full video offline, limiting their applications for scenarios requiring online decisions. Thus, in this work, we explore how to perform \taskshort \textit{online} and \textit{without training}.  
We achieve this by exploiting the zero-shot capabilities of recent Large Multimodal Models (LMMs). In particular, we use LMMs to predict the step associated with a restricted %
set of frames, without access to the whole video. We show that this online strategy without task-specific tuning outperforms offline and training-based models. Motivated by this finding, we develop \methodfull (\methodshort), further injecting knowledge of past frames into the LMM-based predictions. \methodshort exploits Bayesian filtering principles, 
modeling step transitions via (i) a dependency matrix extracted through large language models and (ii) an estimation of step progress.
Experiments on three datasets show superior performance of \methodshort over state-of-the-art training-based offline methods.
\end{abstract}

%% file: sections/1_introduction.tex
\section{Introduction}
\label{sec:intro}

Grounding procedural steps in videos is crucial for enabling machines to follow along and assist humans in complex tasks like cooking a recipe, assembling furniture, or performing maintenance work. 
This ability is particularly valuable for real-time procedural guidance in AR/XR applications, where recognizing task progress allows users wearing headsets or smart glasses to receive timely, step-specific instructions. Specifically, the task of \taskfull(\taskshort) takes as input a list of procedural steps extracted from an instructional article (\eg, a recipe or how-to guide), and a video performing the same task, with the goal of identifying which of the steps are performed in the video.

Existing \taskshort approaches align procedural steps descriptions with their corresponding video frames~\cite{chen2024learning,han2022temporal,li2024multi,mavroudi2023learning}. However, these strategies face two key limitations. First, they need a training set, entailing the cost of collecting (and potentially annotating) it. Moreover, a training set could bias models toward the specific videos and procedural tasks that are depicted, limiting their generalization capability. Second, they assume offline processing, where the entire video is available ahead of time. This makes them unsuitable for real-world applications processing a live video stream.  

\begin{figure}[ht!]
    \centering
    \includegraphics[width=\linewidth]{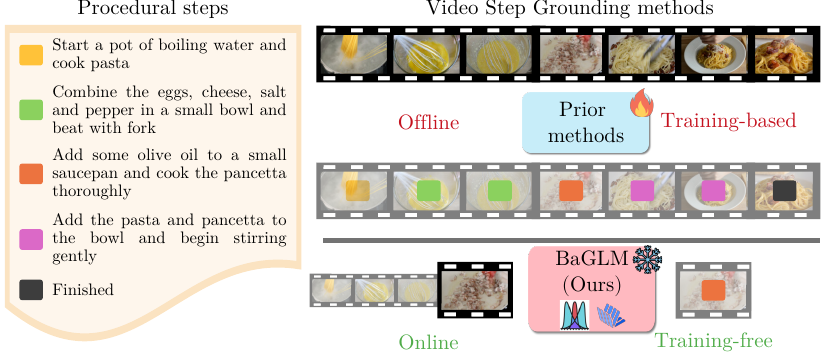}
    \caption{We tackle \textbf{Video Step Grounding} with \textbf{\methodshort}, a  \textit{training-free} approach which combines Bayesian filtering with Large Multimodal Models to enable \textit{online} inference over video streams.%
    }
    \label{fig:teaser}
\end{figure}
To overcome these limitations, we explore how to address  \taskshort \textit{online} and \textit{without training}, \ie, operating on a streaming video (Fig.~\ref{fig:teaser}). This objective is challenging as it requires performing predictions with partial evidence (\ie, having access to only a subset of the video frames) and without the possibility of extracting task-specific representations that would be typically learned from a dedicated training set. In this scenario, we explore the zero-shot capabilities of powerful Large Multimodal Models (LMMs) \cite{bai2025qwen2,li2025llavaonevision,chen2024expanding} for \taskshort. Specifically, given the set of steps, we prompt LMMs to predict the step corresponding to the current video segment. Surprisingly, models such as InternVL \cite{chen2024expanding} could already surpass state-of-the-art training-based methods~\cite{li2024multi,mavroudi2023learning}, despite only having access to a single segment at a time. This highlights a strong potential for addressing \taskshort without training.

Building on the zero-shot capabilities of LMMs, we show how performance can be further improved by modeling temporal dependencies across steps and leveraging past information. To this end, we propose \methodfull{} (\methodshort{}), a training-free approach for \taskshort{} that combines Bayesian probabilistic modeling with LMMs. We harness a Large Language Model (LLM) to estimate a dependency matrix,  capturing whether one step is a prerequisite of another. From this matrix, we compute transition probabilities to each step in the current video segment (\ie, the \textit{predict} step of the Bayesian filter). This prior refines the LMM direct prediction (\ie, the \textit{update} step), injecting past temporal knowledge into the model's output.
The transition model is updated over time, following the progress of each step estimated by the LMM. On three publicly available datasets (HT-Step~\cite{afouras2023ht}, CrossTask~\cite{zhukov2019cross}, Ego4D Goal-Step~\cite{song2023ego4d}), \methodshort consistently outperforms existing methods with significant margins, achieving state-of-the-art results on this challenging task. 

We make four key \textbf{contributions}:
\ding{172} We present the first study to address \taskshort in an online, training-free setting, eliminating the need for data collection and better aligning with practical application needs;
\ding{173} We show that the zero-shot LMMs can surpass specialized, training-based methods, revealing their potential for addressing \taskshort. \ding{174} We propose a method, \methodshort, which incorporates priors from past video frames into LMMs through Bayesian filtering, modeling the temporal dependencies across steps via LLM queries; \ding{175} We extensively evaluate BAGLM on three datasets, showing that it outperforms
state-of-the-art offline methods.

%% file: sections/2_related.tex
\section{Related work}
\label{sec:related}

\textbf{Video Step Grounding.} 
Earlier works in \taskshort adopted weakly-supervised learning approaches. For instance, Zhukov \textit{et al.} \cite{zhukov2019cross} proposed to learn a model from instructional narrations and a list of steps derived from temporal constraints, sharing components across tasks with similar actions or objects. Han \textit{et al.}~\cite{han2022temporal} proposed a co-training framework that combines a Temporal Alignment Network (TAN) with a dual-encoder architecture, predicting step boundaries by aligning videos and narrations, using pseudo-labels derived from cross-modal agreement. VINA~\cite{mavroudi2023learning}
considered step descriptions from WikiHow \cite{wikihow} and learned to temporally ground them in videos without manual supervision. Recent methods have increasingly leveraged language models and large-scale pretraining. 
MPTVA \cite{chen2024learning} introduced a multi-pathway alignment strategy using LLM-filtered narration summaries and multiple sources of alignment, merging them to create robust pseudo-labels for training. NaSVA \cite{li2024multi} addressed multi-sentence grounding by leveraging LLMs to transform noisy ASR transcripts into procedural steps, aligning them with video content using a narration-based similarity score. 

Together, these approaches illustrate the progress from weakly supervised models leveraging narrations to more sophisticated ones that integrate language models, multimodal alignment, and robust pseudo-labeling. However, most methods still rely on extensive training, domain-specific fine-tuning, and access to the full video, assumptions that limit their real-world applicability. To the best of our knowledge, \methodshort is the first training-free online solution that addresses these limitations.

\textbf{Video-Language Alignment} refers to the task of measuring the semantic consistency between a video and its corresponding textual description. Early approaches \cite{hessel2021clipscore,shi2022emscore} tackled this by relying on the cosine similarity between video frames and captions within the embedding space of CLIP \cite{radford2021learning}. However, these methods are inherently limited by the well-known shortcoming of CLIP, \ie, by its inability to effectively capture temporal dynamics in text descriptions. As a result, recent works have shifted toward leveraging LMMs \cite{lin2023revisiting,wu2024towards,wu2024vila,li2024evaluating,zanella2025can} and adopting metrics such as VQAScore \cite{lin2024evaluating}, which are obtained from video question answering to better account for the temporal dimension.

Building on these recent studies, we propose to employ an LMM to assess the alignment between instructional steps and temporal video segments, requiring fine-grained video understanding. VSG is particularly challenging because key steps in instructional tasks often involve similar objects or scenes. For instance, ``inserting a screw'' or ``aligning parts'' may look similar in tasks like ``Assemble a chair'' and ``Fix a table'', making them hard to distinguish without nuanced semantic understanding.

%% file: sections/3_preliminary.tex
\section{On using Large Multimodal Models for Video Step Grounding}\label{sec:preliminary}

Large Multimodal Models are powerful pretrained models that have demonstrated impressive zero-shot performance on a wide variety of tasks without further tuning. As previous solutions for \taskshort are training-based, we wonder whether off-the-shelf LMMs can address \taskshort. This section begins by providing a formal definition of the \taskshort task, clearly distinguishing between its offline and online settings (Sec.~\ref{sec:preliminary:problem}). 
We then present the findings of our preliminary empirical investigation into the use of LMMs for addressing \taskshort, along with key insights gained from this study (Sec.~\ref{sec:method:lmms}).

\subsection{Video Step Grounding}
\label{sec:preliminary:problem}
Given a video of a task composed of a series of actions, \taskfullmin aims to detect which actions (or steps) appear in the video. 
Formally, let us denote the set of steps composing a task as $\steps = \{\step_i \}_{i=1}^{\numsteps}$, where each step $\step_i \in \steps$ is expressed in the natural language space $\textspace$, and $\numsteps$ is the number of steps. Moreover, let us denote with $\video$ a video in the space $\videospace$, split into $\numsegments$ non-overlapping segments $\videosegments = \{\segment_t\}_{t=1}^{\numsegments}$. 
\taskshort aims to identify which steps in $\steps$ are shown in the video. 

The offline setting of \taskshort, as addressed by previous works \cite{li2024multi,mavroudi2023learning}, assumes access to the whole video (\ie, the full set $\videosegments$) when performing the task. In this work, we focus on \textit{online} \taskshort, assuming a video stream where segments arrive %
one after the other, and we perform the task having only access to the current segment and the segments preceding it, \ie, $\videosegments_{1:t} = \{\segment_1, \ldots, \segment_{t}\}$. 
Thus, online \taskshort aims to predict whether a segment $\segment_t\in\videosegments$ depicts a step $\step_i\in\steps$, given only the previous video segments $\videosegments_{1:t}$. While we do not exploit this possibility, a model may also use the current segment to update predictions on past ones, contrary to the single prediction of the offline setting. %

\subsection{Large Multimodal Models are strong baselines for \taskshort}
\label{sec:method:lmms}

Let us define a large multimodal model $f_\mathtt{LMM}$ via three elements: the visual encoder $\visualencoder$, the text encoder $\textencoder$, and a text decoder $\textdecoder$. 
The encoders map their respective inputs into a shared $d$-dimensional embedding space, \ie, $\visualencoder:\videospace \rightarrow \sharedspace$ and $\textencoder:\textspace \rightarrow \sharedspace$. 
The decoder maps the visual and textual inputs into a probability simplex $\wordssimplex$, over the LLM vocabulary $\words$\footnote{For simplicity, we omit the words' tokenization, assuming text prompts and videos are encoded equally.
}, \ie, $\textdecoder:\sharedspace\times\sharedspace \rightarrow \wordssimplex$. The next token is sampled from this probability vector. 

\noindent\textbf{Preliminary experiment.} We propose to frame the problem of \taskshort as a multi-choice question answering task where the LMM, prompted with the current video segment and all possible steps, has to predict as answer either one of them or ``none''. 
Formally, given the current segment $\segment_t$, we prompt the LMM $f_\mathtt{LMM}$ using information regarding the task and step, obtaining the score for a step $\step_i$ as:
\begin{equation}
    \label{eq:lmm-pred}
   f_\mathtt{LMM}(\segment_t, \pi_{\texttt{VSG}})[i] = \textdecoder\left(\visualencoder(\segment_t), \textencoder(\pi_{\texttt{VSG}})\right)[i]
\end{equation}
where $\pi_{\texttt{VSG}}$ is the task prompt (see details in \suppmat) and $\textdecoder(\cdot,\cdot)[i]$ denotes the probability that the next predicted token is $i$ (corresponding to step $\step_i$), normalizing the scores across the multi-choice options. 
Note that, to account for no-step occurring, we include an additional option ``none of the above'' ($\textdecoder(\cdot,\cdot)[\numsteps+1]$). By normalizing the LMMs' probabilities over each choice, we map the segment into a probability simplex over the steps and the ``none'' option, \ie, $f^\mathtt{VSG}_\mathtt{LMM}:\videospace\times\steps\rightarrow\predsimplex$. 

\begin{figure}[t]
    \centering
    \includegraphics[width=\linewidth]{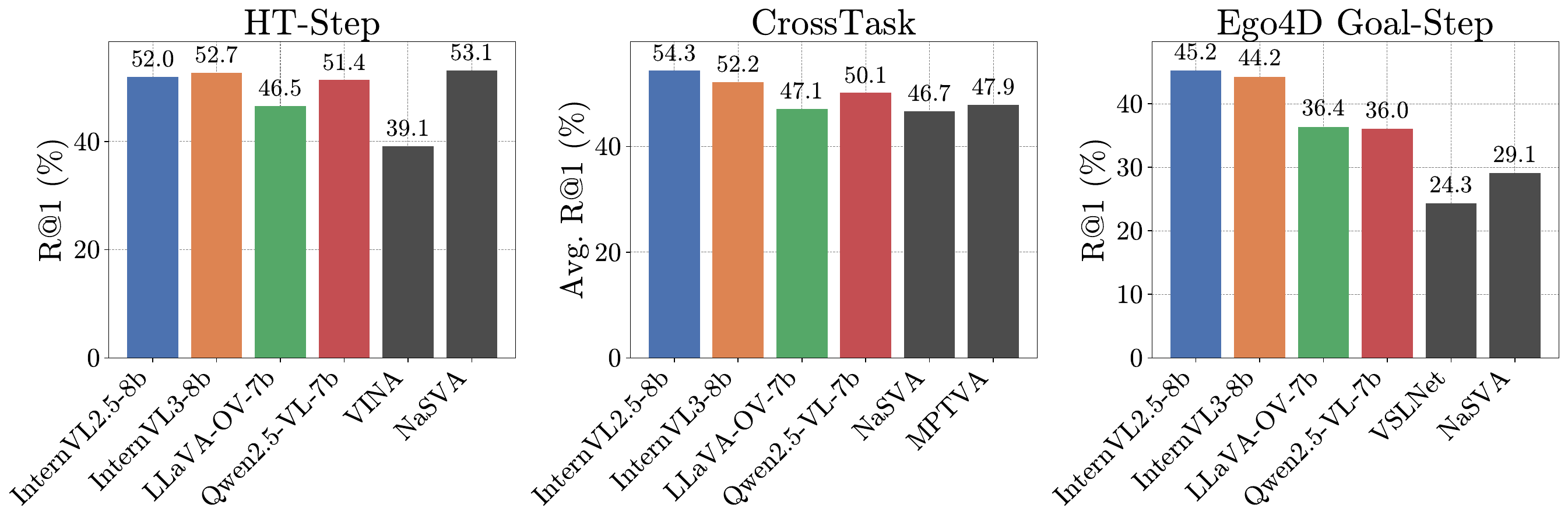}
    \caption{Comparison of \taskshort performance on HT-Step, CrossTask, and Ego4D Goal-Step datasets, prompting LMMs with step options and video segments in an online fashion. For reference, we also show the performance of the top two performing methods from the state of the art (dark bars).}
    \label{fig:preliminary}
\end{figure}

\noindent\textbf{Datasets \& metrics.} 
We evaluate methods on three public datasets: \textit{CrossTask}~\citep{zhukov2019cross}, \textit{HT-Step}~\citep{afouras2023ht}, and \textit{Ego4D Goal-Step}~\citep{song2023ego4d}. \textit{HT-Step} is a benchmark for procedural step grounding~\citep{afouras2023ht}, where the goal is to align steps from an instructional article with an input how-to video. 
The dataset provides two types of test sets: one for seen activities and another for unseen activities. The seen validation and test splits follow \cite{mavroudi2023learning}, each containing 600 videos in total, with 5 videos per activity across 120 activities. 
We evaluate with the validation set of seen classes, as the evaluation server hosting the test sets for the seen and unseen classes is unavailable.

\textit{CrossTask} is an established instructional video benchmark for zero-shot step localization \citep{zhukov2019cross}. It contains about 4.8k instructional videos, covering 18 primary tasks and 65 related tasks. Only videos in the primary tasks are annotated as steps with temporal segments from a predefined taxonomy. We follow the same evaluation set as indicated in~\cite{li2024multi}, using videos from primary tasks.

\textit{Ego4D Goal-Step}~\citep{song2023ego4d}, a subset of the Ego4D~\cite{grauman2022ego4d}, includes 851 videos averaging 26 minutes in length. Unlike CrossTask and HT-Step, Ego4D videos are collected without predefined tasks. Annotators label them hierarchically, first identifying goals (\eg, \textit{Makes the bread}), then steps (\eg, \textit{Prepares the bread}), and finally substeps (\eg, \textit{weigh the dough}). We evaluate on its validation split.

For both HT-Step and CrossTask, we follow the standard evaluation protocol~\citep{li2024multi,chen2024learning}, providing for each video the full set of steps for its task as multiple-choice options in the prompt. On average, each task includes about 10 steps in HT-Step and 7.5 in CrossTask. For Ego4D Goal-Step, we use step-level descriptions only (excluding substeps) and apply text normalization with \texttt{spaCy}\footnote{We use the model available at \url{https://spacy.io/models/en\#en_core_web_sm}}: we lowercase, lemmatize (preserving plural nouns and verbal adjectives), and normalize whitespace and punctuation. After preprocessing, the average number of steps per video is 17.

Following \cite{li2024multi,chen2024learning}, we report Recall@1 (R@1) on HT-Step and Ego4D Goal-Step, measuring whether the top-scoring timestamp for each step falls within the ground-truth interval. For CrossTask, we report Average Recall@1 (Avg.R@1), computed by averaging per-task R@1 across all primary tasks. 

\noindent\textbf{Discussion.} Fig.~\ref{fig:preliminary} shows the results on the three datasets, using four LMMs with strong performance in video understanding benchmarks \cite{plizzari2025omnia}: LLAVA-OneVision-Qwen2-7B \citep{li2025llavaonevision}, Qwen2.5-VL-7B-Instruct \citep{bai2025qwen2}, InternVL2.5-8B \citep{chen2024expanding}, and InternVL3-8B \citep{zhu2025internvl3}. For reference, we also include the top two state-of-the-art methods on each dataset, considering both in-domain ones, \ie, trained and evaluated on the same dataset (VINA \cite{mavroudi2023learning} and NaSVA \cite{li2024multi} on HT-Step, VSLNet \cite{zhang2021natural} on Ego4D Goal-Step) and out-of-domain ones (\ie,  NaSVA on CrossTask and Ego4D Goal-Step, MPTVA~\cite{chen2024learning} on CrossTask). 
Among the LMMs, InternVL2.5-8B scores the best performance on CrossTask and Ego4D Goal-Step, outperforming MPTVA by 6.4\% and NaSVA by 16.1\%, respectively. However, on HT-Step, NaSVA slightly surpasses InternVL2.5-8B (53.1 vs. 52.0). Overall, the LMMs outperform prior methods on CrossTask and Ego4D Goal-Step, while showing comparable performance on HT-Step.

\noindent\textbf{Remarks.} Considering the lack of task-specific tuning, these results demonstrate that zero-shot LMMs perform surprisingly well on \taskshort, accessing only the current segment. 
A natural question is whether we can (i) inject information from past segments into LMM's predictions, refining them, while (ii) maintaining the zero-shot, training-free advantages of LMMs. In the following, we explore how we can achieve this by drawing inspiration from Bayesian filtering principles.

%% file: sections/4_method.tex
\section{\methodfull}
\label{sec:method}

\begin{figure}[t]
    \centering
    \includegraphics[width=\linewidth]{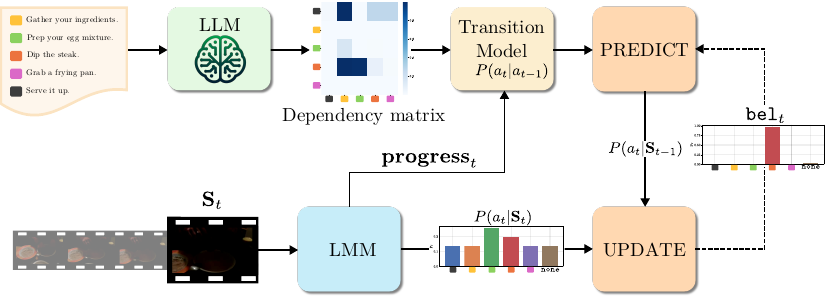}
    \caption{\textbf{Overview of \methodshort.} Given a sequence of steps, an LLM is used to estimate a dependency matrix among them. This matrix is used to compute step transition probabilities employed during the \textbf{predict} step of a Bayesian filter. As the video progresses, the transition model is updated using estimates of each step's progress from an LMM. The \textbf{update} step of the filter merges this with the predictions from the LMM, refining the output.}
    \label{fig:method}
\end{figure}

In Sec.~\ref{sec:preliminary} we have shown that LMMs are effective at \taskshort without any task-specific tuning. However, they act without memory of past knowledge, performing step prediction by only looking at the current segment. 
Therefore, uncertain predictions (\eg, due to the segment acquisition) cannot benefit from past evidence (\eg, step performed in the previous segment), leading to potential model mistakes.

Formally, Eq.~\eqref{eq:lmm-pred} provides an estimate for $P(\stepvar=\step_t|\segment_t)$,  where $\stepvar$ is a discrete random variable taking values in the set $\steps \cup \mathtt{none}$ with $\mathtt{none}$ denoting any action outside the set of steps. 
Differently, we would like to perform step prediction conditioned on all previous segments, estimating $P(\stepvar=\step_t|\videosegments_{[1:t]})$,without the need of storing the whole history. 
Inspired by Bayesian filtering, a probabilistic technique addressing sequence modeling from past observations \cite{chen2003bayesian}, we propose \textit{\methodfull (\methodshort)}. \methodshort (Fig.~\ref{fig:method}) estimates the transition probabilities across steps through the step dependencies estimated by an LLM, to refine LMMs' predictions. 

In the following, we will first present the generic formulation of Bayesian filtering in Sec.~\ref{sec:method:bayesian_filtering}, and how we revisit the predict (Sec.~\ref{sec:method:predict}) and update (Sec.~\ref{sec:method:update}) steps for \taskshort using LMMs and LLMs.

\subsection{Bayesian filtering}
\label{sec:method:bayesian_filtering}

Let $\hiddenstate$ be a state of a process, $\Hiddenstate_t$ be its corresponding random variable at time $t$, and $\measurement_i$ be the observation at time $i\leq t$. The goal of Bayesian filtering \cite{chen2003bayesian}  is to compute the posterior:
\[
\beliefn_t(\hiddenstate) = P(\Hiddenstate_t=\hiddenstate|\measurement_1, \cdots, \measurement_t),
\]
via two essential steps: the \textbf{predict} step first computes a prior over the possible predictions using past estimations; the \textbf{update} step then estimates the prior with the current observations. 
In the following, we will denote $\measurement_{1:t} = \{\measurement_1, \cdots, \measurement_t\}$ for simplicity. 
Using the chain rule, we can write the posterior as:
\begin{equation}
\beliefn_t(\hiddenstate) = P(\Hiddenstate_t=\hiddenstate|\measurement_{1:t})
=  \frac{1}{P(\measurement_t | \measurement_{1:t-1})}\cdot{P(\measurement_t | 
\Hiddenstate_t=\hiddenstate, \measurement_{1:t-1})} \cdot {P(\Hiddenstate_t=\hiddenstate| \measurement_{1:t-1})},
\label{eq:gen_filter}
\end{equation}
where the first term is a normalization factor, the second is the likelihood of the current observation given the past ones and the current state, and the last is the prior over the states from the observations. 

To further simplify Eq.~\eqref{eq:gen_filter}, we consider two \textit{assumptions}: (i) we have a hidden Markov observation model, and thus $P(\measurement_t | 
\Hiddenstate_t=\hiddenstate, \measurement_{1:t-1}) = P(\measurement_t | 
\Hiddenstate_t=\hiddenstate)$; (ii) we have an initial prior over the states independent from the first observation, \ie, $P(\Hiddenstate_0=\hiddenstate | \measurement_1)=P(\Hiddenstate_0=\hiddenstate)$.

Adding the first assumption to the second term and using the Chapman-Kolmogorov equation on $P(\Hiddenstate_t=\hiddenstate| \measurement_{1:t-1})$, we obtain:
\begin{gather}
\beliefn_t(\hiddenstate) 
= \overbrace{\frac{1}{\underbrace{P(\measurement_t | \measurement_{1:t-1})}_{\text{normalization factor}}}\cdot \underbrace{P(\measurement_t | 
\Hiddenstate_t=\hiddenstate)}_{\text{likelihood}}}^{\text{\textbf{update product}}}\cdot \overbrace{\sum_{\hiddenstate_i \in \mathcal{X}} \underbrace{P(\Hiddenstate_t=\hiddenstate | \Hiddenstate_{t-1}=\hiddenstate _i)}_{\text{transition model}}\cdot \underbrace{\beliefn_{t-1}(\hiddenstate_i)}_{\text{accumulated belief}}}^{\text{\textbf{predict step}}}, 
\label{eq:decomposed_gen_filter}
\end{gather}
where $\mathcal{X}$ is the set of possible states. 
The second term corresponds to the \textbf{predict} step, computing an estimate of the current state using the prior belief. The transition model describes the likelihood of a state given the previous one, while accumulated belief denotes the likelihood of the previous state as recursively accumulated via Eq.~\eqref{eq:decomposed_gen_filter}. The first term refers to the \textbf{update} step, where the predicted state probability is multiplied by the likelihood and normalization factor to obtain the final estimate. 

\textbf{A Bayesian filtering view on \taskshort.} To adapt Eq.~\eqref{eq:decomposed_gen_filter} to \taskshort we must  define our states and observations. The state is what we want to estimate, \ie, the step $\step$ in the current segment. The observation is the input we receive from the environment:  the segment $\segment$ itself. Thus, we obtain the update step as:
\begin{equation}
\beliefn_t(\step) = P(\stepvar_t=\step | \videosegments_{1:t})
=  \frac{{P(\segment_t | \stepvar_t =\step, \videosegments_{1:t-1})}}{{P(\segment_t | \videosegments_{1:t-1})}}\cdot \sum_{\step_i\in \steps}{P(\stepvar_t = \step | \stepvar_{t-1} = \step_i)}\cdot \beliefn_{t-1}(\step_i),
\label{eq:vsg_filter}
\end{equation}
where $\stepvar_j$ is a random variable over the possible steps for the $j^{\text{th}}$ segment. We keep the original model's assumptions: an initial prior over steps independent of the first segment, and conditional independence of the current segment given the step (which holds for step-level semantics). In the following, we detail the implementation of each component in Eq.~\eqref{eq:vsg_filter}. 

\subsection{PREDICT: modeling dependencies among steps via language and progress priors}
\label{sec:method:predict}

A peculiarity of \taskshort is that actions depend on each other: these dependencies provide priors on the actions performed in future segments, allowing us to build a transition model, needed in Eq.~\eqref{eq:vsg_filter}.
We exploit the internal knowledge of an LLM to estimate such dependencies. Specifically, we query the LLM to identify when a step must be completed before another can occur (\ie, is a prerequisite). 
As the dependency might be ambiguous, we instruct the LLM to estimate a probability rather than a binary score, resulting in a matrix $\dependencymatrix \in \mathbb{R}^{\numsteps \times \numsteps}$, where each entry $\dependencymatrix_{i,j}$ is the probability that step $\step_j$ is a prerequisite of step $\step_i$. 
We initialize the transition matrix as $\transitionmatrix = \dependencymatrix^\top$, allowing for self-transitions (\ie, $\transitionmatrix_{i,i}=1$), and transitions from all steps to those with no prerequisites (\ie, $\transitionmatrix_{i,j}=1$ if $\sum_{j=1}^\numsteps\transitionmatrix_{i,j}=0$), normalizing $\transitionmatrix$ across rows. 

\textbf{Modeling the action progress.} The transition matrix $\transitionmatrix$ is static and purely based on the task description, without reflecting how the likelihood of a step evolves during the video. For example, if \textit{boil water} is a prerequisite for \textit{cook pasta} and \textit{boil water} has not yet finished, the video cannot transition to \textit{cook pasta}. Instead, it is more likely to continue showing \textit{boil water} or switch to a parallel steps like \textit{prepare the sauce}. Conversely, once \textit{cook pasta} is completed, it becomes unlikely that the video will return to an earlier step as \textit{boil water}. 
We therefore adjust $\transitionmatrix$ accounting for both step dependencies and the estimated step progress, introducing two scores: \textit{readiness} and \textit{validity}. 
Intuitively, a step is ready when its prerequisites are sufficiently complete, while a step is a valid candidate for a segment if its successors have not yet been completed \cite{shen2024progress}.

To achieve this, we query the LMM to infer the execution progress of each step within a video segment.  
Given a segment $\segment$ and the prompt $\pi_{\mathtt{prog}}$, we define the estimated progress for step $\step_i$ as: 
\[
\progress_t[i] = \sum_{j=0}^{9} j\cdot f_\mathtt{LMM}(\segment_t, \pi_{\mathtt{prog}})[j],
\]
where $f_\mathtt{LMM}(\segment_t, \pi_{\mathtt{prog}})[j]$ denotes the model’s output probabilities over the vocabulary. We treat the model’s probability distribution over the tokens $\{0, 1, \ldots, 9\}$, as the distribution over progress levels. From the latter and $\dependencymatrix$, we can compute whether an action is ready to be performed.

\noindent\textbf{\textit{Step readiness.}} For each step $\step_i$, we compute readiness as the weighted maximum progress of its prerequisite steps, \ie, 
\begin{equation}
\readiness_t[i] = \frac{\sum_{j=1}^{\numsteps} \dependencymatrix_{i,j} \cdot \max_{\tau<t} \progress_\tau[j]}{\sum_{j=1}^{\numsteps} \dependencymatrix_{i,j}},
\label{eq:readiness}
\end{equation}
where we measure the progress of a step as its maximum value across all preceding segments (\ie, $\tau < t$), performing a weighted average across all steps. With Eq.~\eqref{eq:readiness}, we get high values in case all predecessors of an action have a high progress, and low otherwise. 

\noindent\textbf{\textit{Step validity.}} Contrary to the readiness, the step validity is given by:

\begin{equation}
\validity_t[i] = \frac{\sum_{j=1}^{\numsteps} \dependencymatrix_{j,i} \cdot (1 - \max_{\tau<t} \progress_\tau[j])}{\sum_{j=1}^{\numsteps} \dependencymatrix_{j,i}}.
\label{eq:validity}
\end{equation}
This value is high when no successors of $\step_i$ in $\dependencymatrix$ have been executed yet, and low otherwise. 
Finally, we adjust the transition matrix $\transitionmatrix$ by accounting for readiness and validity of each step, \ie, 

\begin{equation}
\tilde{\transitionmatrix}_t[i, j] = \frac{\transitionmatrix[i, j] \cdot \readiness_t[j] \cdot \validity_t[j]}{\sum_{k=1}^{\numsteps}\transitionmatrix[i, k] \cdot \readiness_t[k] \cdot \validity_t[k]}.
\label{eq:transition}
\end{equation}

\textbf{Predict step.} Exploiting the transition model of Eq.~\eqref{eq:transition}, 
the predict step of Eq.~\eqref{eq:vsg_filter} becomes:
\begin{equation}
   \mathtt{predict}_t(\step_i)= \sum_{\step_j\in \steps}\tilde{\transitionmatrix}_t[j, i]\cdot \beliefn_{t-1}(\step_j).
   \label{eq:vsg_filter_predict}
\end{equation}

\subsection{UPDATE: using LMM estimates to re-weigh the belief over steps}
\label{sec:method:update}

In the \textbf{update step}, we multiply the step prior for the observation likelihood $P(\segment_t | \stepvar_t =\step)$ and a normalization factor independent from the steps. However, due the cardinality of $\videospace$, computing $P(\segment_t | \step_t, \videosegments_{t})$ is intractable. On the other hand, we follow Eq.~\eqref{eq:lmm-pred}, estimating $(\segment_t | \stepvar_t =\step)$ directly from the LMM. 
Formally, we use the Bayes rule and write: 
\begin{equation}
P(\segment_t | \stepvar_t = \step_i) = \frac{P(\segment_t)}{P(\stepvar_t = \step)}\cdot P(\stepvar_t = \step_i|\segment_t) = \frac{P(\segment_t)}{P(\stepvar_t = \step_i)}\cdot f_\mathtt{LMM}(\segment_t, \pi_{\texttt{VSG}})[i],
\label{eq:likelihood}
\end{equation}
where $P(\segment_t)$ is a prior on the segments independent from the steps, $P(\stepvar_t = \step_i)$ is a prior over the steps independent from the observation. We replace $P(\stepvar_t =\step|\segment_t)$ with the LMM prediction. 
While for $P(\stepvar_t = \step_i)$ there are various possible choices, in our approach, we consider the prior to be uniform, ablating this choice in the \suppmat. 

\textbf{Final filtering model.} 
Considering the uniform prior over the states and merging Eq.~\eqref{eq:vsg_filter_predict} into Eq.~\eqref{eq:likelihood}, we obtain the final belief $\beliefn_t(\step_i)$ for step $\step_i$ and segment $\segment_t$ as:
\begin{equation}
    \beliefn_{t}(\step_i) = \frac{1}{\mathcal{Z}} \cdot {f_\mathtt{LMM}(\segment_t, \pi_{\texttt{VSG}})[i]}{\mathcal{Z}}\cdot \sum_{\step_j\in \steps}\tilde{\transitionmatrix}_t[j, i]\cdot \beliefn_{t-1}(\step_j),
    \label{eq:final_model}
\end{equation}
where $\mathcal{Z}$ is a normalization factor containing all elements independent of the specific step $\step_i$.

%% file: sections/5_experiments.tex
\section{Experiments}
\label{sec:exp}
In this section, we describe our experimental protocol and present the comparison \wrt the state of the art (Sec.~\ref{sec:exp:comparison}). Finally, we perform a detailed study on \methodshort (Sec.~\ref{sec:exp:ablation}). We use the same datasets and metrics described in Sec.~\ref{sec:preliminary} in our experiments.

\noindent\textbf{Implementation details.}
Our method is implemented considering InternVL2.5-8B \citep{chen2024expanding} as our LMM, based on the results of Sec.~\ref{sec:preliminary}. We employ LLaMA3-70B-Instruct \cite{grattafiori2024llama} as our LLM of choice to derive our transition model. To test our model, we split videos into sequences of non-overlapping 2-second segments, providing them as input to the LMM one after the other.  
We ran all experiments on a single NVIDIA H100 64GB GPU, except for LLaMA3-70B-Instruct \cite{grattafiori2024llama}, which required 4 H100 GPUs.

\noindent\textbf{Baselines.} We compare our method with several state-of-the-art approaches for \taskshort: Zhukov \textit{et al.}~\citep{zhukov2019cross}, HT100M~\citep{miech2019howto100m}, VideoCLIP~\citep{xu2021videoclip}, MCN~\citep{chen2021multimodal}, DWSA~\citep{shen2021learning}, MIL-NCE~\citep{miech2020end}, VT-TWINS~\citep{ko2022video}, UniVL~\citep{luo2020univl}, VINA~\citep{mavroudi2023learning}, TAN*~\citep{han2022temporal,afouras2023ht}, NaSVA~\citep{li2024multi}, and MPTVA~\citep{chen2024learning}. We also implement an online variant of NaSVA~\citep{li2024multi}, which introduces causal masking in the transformer encoder's self-attention layers to restrict attention to past segments only. Reported results are taken from the original papers, except for NaSVA on HT-Step and Ego4D Goal-Step, and VSLNet~\citep{zhang2021natural} on Ego4D Goal-Step, where we use the authors' released code. These are marked with $\dagger$ in the tables. 

All baselines are training-based and offline, except for our online variant of NaSVA, and use HowTo100M~\cite{miech2019howto100m} as training set or pre-training, except for VSLNet (trained on Ego4D Goal-Step), for Zhukov \textit{et al.} \cite{zhukov2019cross} and DSWA~\citep{shen2021learning} (trained on CrossTask) and for MPTVA~\citep{chen2024learning} (trained on a subset of HowTo100M). For CrossTask, VideoCLIP~\cite{xu2021videoclip} and UniVL~\cite{luo2020univl} perform a further fine-tuning step on CrossTask data.

\subsection{Comparison with state-of-the-art methods}\label{sec:exp:comparison}
Tab.~\ref{tab:htstep_ego4d_goalstep} and Tab.~\ref{tab:crosstask} report the results of our evaluation in the three considered datasets. 
Overall, \methodshort, built on top of InternVL2.5-8B, outperforms all offline methods that rely on noisy supervision from narrations without manual annotations. Notably, it outperforms the current state-of-the-art method, NaSVA, by 4.3\% on HT-Step, which consists of videos from HowTo100M, the same dataset used for NaSVA's self-training.
The improvements of \methodshort over NaSVA are especially significant on CrossTask and Ego4D Goal-Step, with gains of 13.1\% and 14.2\%, respectively. The margins in CrossTask are remarkable as there exist methods (\ie, VideoCLIP, UniVL) that are specifically fine-tuned for this domain.
The same applies to Ego4D Goal-Step. This dataset is particularly challenging for models trained on HowTo100M due to the distribution shift introduced by its egocentric videos. On this dataset, \methodshort also outperforms VSLNet by a significant margin (+19\%), despite VSLNet being trained on data from the same domain. 
Notably, all these results have been achieved in an online setting, with all competitors having access to the whole video, contrary to \methodshort. When evaluated under the same online setting, NaSVA's performance drops to 46.1 R@1 on HT-Step and 24.2 on Ego4D Goal-Step (-7\% and -4.9\% compared to its offline variant), remaining well below \methodshort (-11.3\% and -19.1\%). These results further highlight the effectiveness of our approach in the challenging online scenario.

\input{tables/comparison}

\input{tables/ablation}

\begin{figure}[t]
    \centering
    \includegraphics[width=\linewidth]{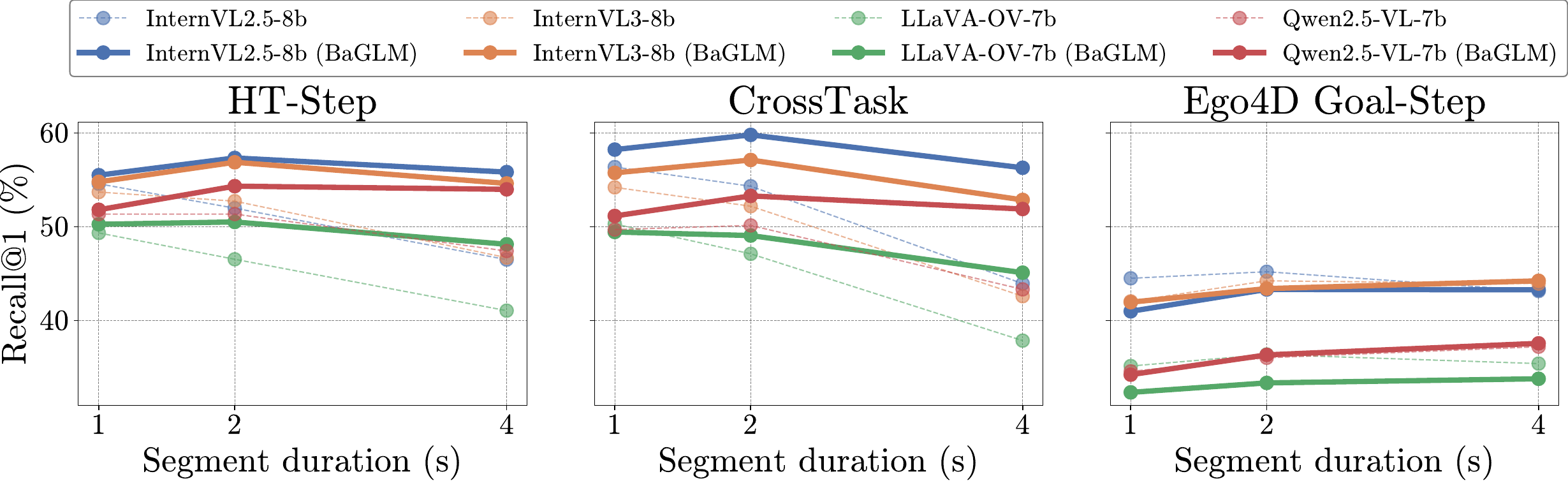}
    \caption{Ablation study on varying the segment duration and on the used LMM.}
    \label{fig:internvl_baglm_segment_duration}
\end{figure}

\input{tables/oracle}

\subsection{Ablation studies}\label{sec:exp:ablation}

In this section, we analyze the key components of \methodshort, evaluating different configurations of the transition model, different LLMs for generating the dependency matrix, and experiments with oracle step dependencies and progress. Additional design choices are discussed in the Appendix.

\textbf{Transition Model.} The transition model estimates the conditional probability of moving from one step to the next and is used in the predict step of Bayesian filtering, as per Eq.~\eqref{eq:vsg_filter_predict}. We first evaluate a static transition matrix, initialized from the dependency matrix. We then study the effect of including the readiness score (\ie, if prerequisites are not completed, Eq.~\eqref{eq:readiness}), the validity one (\ie, if the step should not be re-executed, Eq.~\eqref{eq:validity}), and both. 
As shown in Tab.~\ref{tab:transition_model}, readiness improves the static model by 1.1\% and 0.8\% on two datasets and performs similarly to the static model on the third (-0.1\%). Thus, accounting for the dependency matrix (and if prerequisites are met) tends to improve performance by reducing the score of non-executable actions.

Validity improves the static model by +0.5\%, +0.8\%, and +1.0\% across the three datasets. By considering the status of actions that occur at a later time, it influences the predictions of their prerequisites (\eg, \textit{turn on the stove}) and prevents them from being repeated unnecessarily.

The best results come from combining both, improving performance by 1.5\%, 1.8\%, and 1.2\%. This highlights the benefit of including both types of estimated priors when updating the transition model.

\textbf{Choice of LLMs.} Tab.~\ref{tab:llm} analyzes how \methodshort is affected by the LLM used to generate dependencies between steps. We compare LLaMA-3.3-70B-Instruct~\cite{grattafiori2024llama} and GPT-4.1-mini~\cite{gpt4.1-mini}. The two models achieve comparable overall performance: LLaMA-3.3-70B performs better on HT-Step (+0.3\%) and Ego4D GoalStep (+2.7\%) but worse on CrossTask (-1.1\%). This trend suggests that LLaMA-3.3-70B performs better on datasets with more generic step descriptions, whereas proprietary models like GPT-4.1-mini are more effective at capturing dependencies among more atomic actions.

\textbf{LMMs and segment duration.} In Fig.~\ref{fig:internvl_baglm_segment_duration}, we show how \methodshort's performance varies \wrt the segment duration (from 1 to 4 seconds), considering different LMMs: InternVL2.5 8B~\cite{chen2024expanding}, InternVL3 8B \citep{zhu2025internvl3}, LLaVA-OneVision 7B~\cite{li2025llavaonevision}, and Qwen2.5 7B~\cite{bai2025qwen2}. 

From the figure, we can see three trends. First, \methodshort consistently improves the performance of the underlying LMM it is applied to across all segment durations on both HT-Step and CrossTask, with gains becoming more pronounced as segment duration increases. This is related to the second trend, the impact of the segment duration. Segments that are too short may lack sufficient visual cues to evaluate the video-step alignment or the progress, while longer ones may span over multiple actions, making it harder to localize steps precisely. Using 2-second segments offers the best trade-off, improving performance by +1.9\%, +1.6\%, +0.3\%, and +2.3\% across the three datasets compared to 1-second segments. 

Third, \methodshort does not provide clear advantages on Ego4D Goal-Step (\ie, -1.9\%, -0.8\%, -3\%, and +0.3\%). This can be attributed to challenges specific to this dataset. Ego4D Goal-Step videos are significantly longer on average (28 minutes vs. 6 minutes and less than 5 minutes for HT-Step and CrossTask, respectively) and contain coarser step annotations (53 seconds per segment on average vs. 16 seconds and 10.7 seconds for HT-Step and CrossTask, respectively). These longer videos and broader annotations result in more generic step descriptions, which in turn make it harder to infer accurate dependencies between steps and to estimate progress, as we will analyze in the following. 

\textbf{Use of annotated step boundaries from datasets.} Given the different impact of \methodshort across datasets, we analyze the performance of the Bayesian filtering formulation without potential noise from the dependency matrix and/or the estimation of step progresses. With this aim, we conduct an analysis similar to Tab.~\ref{tab:transition_model}, this time using step dependencies and action progress derived directly from the original datasets. For each video, we construct a chain of steps based on its ground truth temporal order. Note that while these dependencies reflect the observed execution order, \textit{they are not} true semantic constraints, as the same steps may occur in different orders in other videos. Progress is computed as the normalized fraction of completion between a step's start and end timestamps. 

Results of this oracle experiment are presented in Tab.~\ref{tab:oracle}, showing consistent gains across all datasets (\eg, +5.2\% on HT-Step), even the challenging Ego4D Goal-Step (\ie, +38.9\% ). This confirms that the Bayesian filtering formulation is effective, and that jointly improving the elements used to estimate the transition matrix (\ie, progress, dependency) would further boost the results of \methodshort.

%% file: tables/comparison.tex
\begin{table}[t]
  \caption{Comparison between state-of-the-art \inlineColorbox{gray!10}{offline} methods and our \inlineColorbox{DrawioBlue}{online} method \methodshort.
  }
  \centering
  \begin{subtable}[t]{0.52\linewidth}
    \centering
    \input{tables/htstep_ego4d_goalstep}
  \end{subtable}
  \hfill
  \begin{subtable}[t]{0.44\linewidth}
    \centering
    \input{tables/crosstask}

  \end{subtable}
  \label{tab:comparison}
\end{table}

%% file: tables/htstep_ego4d_goalstep.tex
\centering
\caption{HT-Step and Ego4D Goal-Step}
\label{tab:htstep_ego4d_goalstep}
\resizebox{0.9\textwidth}{!}{%
\begin{tabular}{lcc}
\toprule
 & \textbf{HT-Step} & \textbf{Ego4D Goal-Step} \\ 
\textbf{Method} & \textbf{$\uparrow$ R@1 } & \textbf{$\uparrow$ R@1} \\
\midrule
\rowcolor{gray!10}\textit{Offline}                 &                      &                          \\
\rowcolor{gray!10}\textsc{VSLNet} \citep{zhang2021natural}  & -                    & 24.3$^\dagger$           \\
\rowcolor{gray!10}\textsc{TAN*} \citep{han2022temporal}     & 30.7                 & -                        \\
\rowcolor{gray!10}\textsc{VINA} \cite{mavroudi2023learning} & 39.1                 & -                        \\
\rowcolor{gray!10}\textsc{NaSVA} \citep{li2024multi}        & 53.1$^\dagger$       & 29.1$^\dagger$           \\ \midrule
\rowcolor{DrawioBlue} \textit{Online}                  &                      &                          \\
\rowcolor{DrawioBlue}\textsc{NaSVA} \citep{li2024multi}                   & 46.1$^\dagger$        & 24.2$^\dagger$            \\
\rowcolor{DrawioBlue}\textbf{\methodshort}                    & \textbf{57.4}        & \textbf{43.3}            \\ \bottomrule
\end{tabular}%
}

%% file: tables/crosstask.tex
\centering
\caption{CrossTask}
\label{tab:crosstask}
\resizebox{0.7\textwidth}{!}{%
\begin{tabular}{@{}lc@{}}
\toprule
\textbf{Method} & \textbf{$\uparrow$ Avg. R@1 } \\ \midrule
\rowcolor{gray!10}\textit{Offline}                                  &                                   \\                        
\rowcolor{gray!10}Zhukov \textit{et al.} \citep{zhukov2019cross}     & 22.4                              \\
\rowcolor{gray!10}\textsc{HT100M} \citep{miech2019howto100m}                 & 33.6                              \\
\rowcolor{gray!10}\textsc{VideoCLIP} \citep{xu2021videoclip}                 & 33.9                              \\
\rowcolor{gray!10}\textsc{MCN} \citep{chen2021multimodal}                    & 35.1                              \\
\rowcolor{gray!10}\textsc{DWSA} \citep{shen2021learning}                     & 35.3                              \\
\rowcolor{gray!10}\textsc{MIL-NCE} \citep{miech2020end}                      & 40.5                              \\
\rowcolor{gray!10}\textsc{VT-TWINS} \citep{ko2022video}                      & 40.7                              \\
\rowcolor{gray!10}\textsc{UniVL} \citep{luo2020univl}                        & 42.0                              \\
\rowcolor{gray!10}\textsc{VINA} \cite{mavroudi2023learning}                  & 44.8                              \\
\rowcolor{gray!10}\textsc{NaSVA} \citep{li2024multi}                         & 46.7                              \\
\rowcolor{gray!10}\textsc{MPTVA} \citep{chen2024learning}                    & 47.9                              \\ \midrule
\rowcolor{DrawioBlue}\textit{Online}                                   &                                   \\
\rowcolor{DrawioBlue}\textbf{\methodshort}                             & \textbf{59.8}                     \\ \bottomrule
\end{tabular}%
}

%% file: tables/ablation.tex
\begin{table}[t]
\centering
\begin{minipage}[t]{0.52\textwidth}
  \centering
  \caption{Ablation study on the transition model.}
  \resizebox{\linewidth}{!}{%
    \input{tables/ablation_transition_model}

  }
\end{minipage}%
\hfill
\begin{minipage}[t]{0.44\textwidth}
  \centering
  \caption{Ablation study on varying the LLM.}
  \resizebox{\linewidth}{!}{%
    \input{tables/ablation_llm}
  }
\end{minipage}
\end{table}

%% file: tables/ablation_transition_model.tex
\centering
\label{tab:transition_model}
\resizebox{0.9\textwidth}{!}{%
\begin{tabular}{@{}ccccc@{}}
\toprule
\textbf{Readiness} & \textbf{Validity} & \textbf{HT-Step} & \textbf{CrossTask} & \textbf{Ego4D} \\ \midrule
\rowcolor{gray!10} &            & 55.9 & 58.0 & 42.1 \\
\checkmark         &            & 57.0 & 58.8 & 42.0 \\
\rowcolor{gray!10} & \checkmark & 56.4 & 58.8 & 43.1 \\
\checkmark         & \checkmark             & \textbf{57.4}    & \textbf{59.8}      & \textbf{43.3}  \\ \bottomrule
\end{tabular}%
}

%% file: tables/ablation_llm.tex
\centering
\label{tab:llm}
\resizebox{0.9\textwidth}{!}{%
\begin{tabular}{@{}lcc@{}}
\toprule
\textbf{Dataset} & \textbf{LLaMA-3.3-70B} & \textbf{GPT-4.1-mini} \\ \midrule
\rowcolor{gray!10}HT-Step     & \textbf{57.4} & 57.1 \\
CrossTask   & 59.8 & \textbf{60.9} \\
\rowcolor{gray!10}Ego4D       & \textbf{43.3} & 40.6 \\ \bottomrule
\end{tabular}%
}

%% file: tables/oracle.tex
\begin{table}[t]
\centering
\caption{Results with oracle dependencies and step progress.}
\label{tab:oracle}
\resizebox{0.65\textwidth}{!}{
\begin{tabular}{@{}ccccc@{}}
\toprule
\textbf{Progress Oracle} & \textbf{Dep. Matrix Oracle} & \textbf{HT-Step} & \textbf{CrossTask} & \textbf{Ego4D} \\ \midrule
\rowcolor{gray!10} &            & 57.4          & 59.8          & 43.3          \\
\checkmark         &            & 44.9          & 38.6          & 36.0          \\
\rowcolor{gray!10} & \checkmark & 54.6          & 58.3          & 45.2          \\
\checkmark         & \checkmark & \textbf{62.6} & \textbf{66.9} & \textbf{82.2} \\ \bottomrule
\end{tabular}%
}
\end{table}

%% file: sections/6_conclusion.tex
\section{Conclusion}\label{sec:conclusion}
In this work, we introduced \methodshort, a novel training-free approach for online video step grounding. \methodshort uses Bayesian filtering to integrate information from past video frames into LMM predictions. It consists of two key components: a transition model, initialized from step dependency matrices generated by LLMs and updated over time using action progress and dependency constraints; and an observation model, implemented as an LMM, which refines predictions as the video unfolds. We evaluated \methodshort on HT-Step, CrossTask, and Ego4D Goal-Step, showing that it outperforms training-based methods without requiring additional training or data collection. 

\noindent\textbf{Limitations.} We identify two main limitations of our approach. First, \methodshort fully relies on a pre-trained LLM for estimating the step dependency matrix, so its effectiveness is closely tied to the quality of the model predictions. As more advanced LLMs become available, we expect that \methodshort performance will improve accordingly. 
Second, prior estimation within our Bayesian framework may be improved with more specific task knowledge. For example, we estimate step progress by querying an LMM: future works could refine this approach by incorporating priors on step durations.

\noindent\textbf{Acknowledgements.}
This work was supported by the \textit{Ministero delle Imprese e del Made in Italy} (IPCEI Cloud DM 27 giugno 2022 – IPCEI-CL-0000007) and by the European Union (Next Generation EU). Additional support was provided by the EU projects \textit{ELIAS} (No.~101120237) and \textit{ELLIOT} (No.~101214398).
We acknowledge ISCRA for awarding this project access to the LEONARDO supercomputer, owned by the EuroHPC Joint Undertaking, hosted by CINECA (Italy), as well as the EuroHPC Joint Undertaking for awarding us access to MareNostrum5 at BSC, Spain.
The author also thanks Andrea Pilzer for valuable discussions on efficient video decoding.

%% file: appendix/a_prompts.tex
\section{Prompts}
\label{sec:app:prompts}

We use three types of prompts in our approach. For each, the Large Multimodal Model (LMM) is prompted to generate an answer by selecting from a list of options. We then compute the model’s score for each choice by applying a softmax to the logits of the first generated token and taking the probability assigned to the first token of each option. Video frames are positioned relative to the text according to the input formatting required by the underlying LMM.

The task prompt $\pi_{\texttt{VSG}}$ is used to estimate the probability that a given video segment corresponds to each step in the task, or to the ``none'' class. The LMM is prompted with the current video segment, the task \texttt{goal} (\ie, the procedural task), and all candidate \texttt{steps}, and must select either one of the steps or the ``none'' option as the answer.

\begin{tcolorbox}[breakable, enhanced jigsaw, title=$\pi_{\texttt{VSG}}$]
You are watching a video segment of someone attempting to \texttt{\{goal\}}.\\[1ex]
What is the main action being performed in this exact moment?\\
\\
Options:\\
\texttt{\{steps\}}\\
\\
Answer with only the letter label of the correct option (e.g., A, B, ..., Z, AA, AB, etc.), with no extra text.
\end{tcolorbox}

The progress prompt $\pi_{\mathtt{prog}}$ is used to estimate the execution progress of a specific step within a video segment. The LMM is prompted with the current video segment, the task \texttt{goal}, and a given \texttt{step}, and must select a progress value from 0 to 9. Here, 0 indicates that the action is not occurring, 1 means the action is about to begin, 5 corresponds to the middle of the action, and 9 means the action has just finished.

\begin{tcolorbox}[breakable, enhanced jigsaw, title=$\pi_{\mathtt{prog}}$]
You are watching a short video clip.\\
\\
The goal of the person in the video is: \texttt{\{goal\}}\\[1ex]
The specific action of interest is: \texttt{\{step\}}\\
\\
Rate how far along this action is in terms of execution progress, using a scale from 0 to 9:\\
- 0 = the action is not present in this clip\\
- 1 = the action is just beginning or about to begin\\
- 5 = the action is halfway complete\\
- 9 = the action is just finishing or about to finish\\
\\
Respond with a single number from 0 to 9. Do not include any other text.
\end{tcolorbox}

Finally, the prerequisite prompt $\pi_{\mathtt{prereq}}$ is used to estimate step dependencies within a task. The LLM is prompted with the task \texttt{goal}, a \texttt{step}, and a candidate \texttt{prerequisite}, and must answer either ``Yes'' or ``No''.

\begin{tcolorbox}[breakable, enhanced jigsaw, title=$\pi_{\mathtt{prereq}}$]
You are performing the task: \texttt{\{goal\}}\\[1ex]

Is the following step strictly required before another?\\[1ex]

Prerequisite candidate: \texttt{\{prerequisite\}}\\
Target step: \texttt{\{step\}}\\[1ex]

Answer ``Yes'' if the target step cannot be completed correctly without first completing the prerequisite step. Otherwise, answer ``No''.\\[1ex]

Answer:
\end{tcolorbox}

%% file: appendix/b_additional_analyses.tex
\section{Additional analyses}
\label{sec:app:ablation}

In this section, we present additional analyses of \methodshort, including ablations on the prompts used to compute \taskfull scores, the choice of step priors, and step localization performance. We also evaluate generalization on the COIN dataset, analyze the robustness of LLM-generated dependency matrices on HT-Step, and study step prediction performance by temporal position and by the number of candidate steps.

\input{tables/ablation_prompt}

\noindent\textbf{Binary vs multi-choice \taskshort prompt.}
While we compute \taskshort scores by framing the problem as a multi-choice question answering task, an alternative is to treat it as a binary question answering task. In this case, we can ask the LMM whether a given video segment corresponds to a specific step, querying the LMM for each step independently. We compare these approaches by evaluating the off-the-shelf model InternVL2.5-8B using the multi-choice prompt $\pi_{\texttt{VSG}}$ and the binary prompt $\pi_{\texttt{VSG}}^{\texttt{bin}}$:
\begin{tcolorbox}[breakable, enhanced jigsaw, title=$\pi_{\texttt{VSG}}^{\texttt{bin}}$]
You are watching a video segment of someone attempting to \texttt{\{goal\}}.\\[1ex]
Is the person currently performing the action: "\texttt{\{step\}}"?\\
\\
Answer with ``Yes'' or ``No'' only.
\end{tcolorbox}

As shown in Table~\ref{tab:prompt}, using the multi-choice formulation improves performance by 3.4\% on Ego4D Goal-Step. On HT-Step, performance is comparable (a slight drop of 0.5\%), while on CrossTask, the binary formulation performs better by 1.7\%. However, the binary formulation requires querying the LMM $\numsteps$ times per video segment, once per candidate step, whereas the multi-choice formulation requires only one query per segment.

For example, for the HT-Step video \texttt{-25-1nbOki0}, which contains 283 segments (each 2 seconds long) and 10 steps, using $\pi_{\texttt{VSG}}$ requires 24.79 seconds in total (approximately 0.0876 seconds per segment), while $\pi_{\texttt{VSG}}^{\texttt{bin}}$ takes 214.04 seconds (about 0.756 seconds per segment) with one NVIDIA H100 GPU with 64GB of memory. Given the comparable accuracy and significantly lower inference time, we adopt $\pi_{\texttt{VSG}}$ as our default prompt for computing \taskfull scores.

\input{tables/ablation_prior} 

\noindent\textbf{Prior over the states $P(\stepvar_t = \step_i)$.} 
In Eq.~(9), describing the update step, we normalize the scores using a general prior $P(\stepvar_t = \step_i)$ over all possible states. 
In this section, we explore alternatives to the adopted uniform prior in our method. 

Table~\ref{tab:prior} shows the results of our analyses. Row (1) reports results using a context-dependent prior, where the LMM is prompted with $\pi_{\mathtt{next}}$ to predict the probability distribution over the next most likely action given the current segment, the task \texttt{goal}, and all candidate \texttt{steps}:
\begin{tcolorbox}[breakable, enhanced jigsaw, title=$\pi_{\mathtt{next}}$]
You are watching a video segment of someone attempting to \texttt{\{goal\}}.\\[1ex]
What is the most likely next step in the sequence?\\
\\
Options:\\
\texttt{\{steps\}}\\
\\
Answer with only the letter label of the correct option (e.g., A, B, ..., Z, AA, AB, etc.), with no extra text.
\end{tcolorbox} 

Predicting the next action is often similar to predicting the current one, resulting in a prior distribution close to the \taskshort scores. As a result, the probability associated with some true positive steps will be erroneously lowered in favor of the ``none'' class.

Row (2) presents the results of using a prior initialized from a uniform distribution and updated at each timestamp by applying the state transition model. In this prior, transitions to ``none'' are often the most likely as they are not penalized by step validity or readiness. This lowers the likelihood of the ``none'' class in the posterior, promoting false positive predictions. 

Row (3) shows the performance using a uniform prior (our adopted one), which assigns equal probability to all steps. This setting achieves the best results among all the considered prior options.

\input{tables/htstep_localization}
\noindent\textbf{Step localization.}
While we report performance using Recall@1 and Avg. Recall@1 as our main metrics in the main manuscript, following the evaluation protocol from \citep{li2024multi, chen2024learning}, we also evaluate step localization performance (\ie, predicting the start and end timestamps of steps) by leveraging the final belief distribution for each video segment. This results in an alignment matrix of size $\numsegments \times \numsteps$, where $\numsegments$ is the number of video segments and $\numsteps$ is the number of task steps. Following \citep{mavroudi2023learning}, we extract temporal segments from this matrix using a 1D blob detection approach. Specifically, we apply Laplacian of Gaussian (LoG) filters at 13 different scales, corresponding to Gaussian standard deviations ranging from 1 to 480. The confidence score for each predicted segment is computed as the average step probability across all video segments within the predicted temporal boundaries. 
We compare our method against NaSVA \cite{li2024multi}, the best method (see Tab.~\ref{tab:comparison} in the main manuscript) with publicly available code. NaSVA uses a transformer-based architecture where steps act as queries that iteratively attend to video features, producing an alignment matrix between steps and video segments. To ensure a fair comparison, we derive the start and end timestamps for each step by applying the same 1D blob detection procedure used in our method to NaSVA's similarity matrix, computed from its final transformer layer. Before applying blob detection, we convert the similarity scores for each segment into a valid probability distribution over steps by applying a softmax function.

We conduct this analysis on the HT-Step dataset \cite{afouras2023ht} and report results in Tab.~\ref{tab:htstep_localization}. We evaluate localization performance using the HT-Step protocol, which measures article-grounding mean Average Precision (mAP) at various IoU thresholds. This involves treating each step as a class-agnostic text query, computing AP per activity, and averaging the results to obtain overall mAP. \methodshort performs better than NaSVA at all IoU thresholds (\ie, +2.2\%, +0.7\%, and +0.1\%). 

\input{tables/ablation_lmm_localization}

We complement the results of Fig.~\ref{fig:internvl_baglm_segment_duration} in the main manuscript by evaluating how \methodshort performs when applied to different LMMs, \ie, InternVL2.5 8B~\cite{chen2024expanding}, LLaVA-OneVision 7B~\cite{li2025llavaonevision}, Qwen2.5 7B~\cite{bai2025qwen2}, and InternVL3 8B~\cite{zhu2025internvl3}, using mAP as the evaluation metric. From Tab.~\ref{tab:off_the_shelf_vs_baglm_localization}, we can see that adding \methodshort to any LMM consistently improves performance. The improvements are largest at lower IoU thresholds (\eg, +1.9\% at 0.3 \wrt the off-the-shelf InternVL3 8B) and gradually diminish at higher thresholds (\eg, +0.2\% at 0.7).

\input{tables/coin}
\noindent\textbf{COIN.} To further demonstrate the generalizability of \methodshort, we evaluate it on the COIN \cite{tang2019coin} dataset, which contains instructional YouTube videos spanning various domains. We use the test split, consisting of 2,797 videos, of which 2,354 were successfully downloaded (the remainder being unavailable or set to private). These videos cover 12 domains (\ie, Dish, Drink and Snack, Electrical Appliance, Furniture and Decoration, Gadgets, Housework, Leisure and Performance, Nursing and Care, Pets and Fruit, Science and Craft, Sport, and Vehicle) and are categorized into 180 tasks. 

We compare \methodshort against three baselines: (1) InternVL2.5-8B \cite{chen2024expanding} (off-the-shelf), (2) NaSVA \citep{li2024multi} (best offline competitor), (3) an online version of NaSVA. As shown in Tab.~\ref{tab:coin}, BaGLM outperforms NaSVA by 14.2 (offline) and 18.9 (online), and achieves a gain of 3.1 over InternVL2.5-8B. These results suggest that our method generalizes well across diverse domains and tasks.

\input{tables/htstep_prereq_robustness}
\noindent\textbf{Dependency matrix robustness.} We evaluate the robustness of LLM-generated dependency matrices on the HT-Step dataset, which includes 120 cooking-related tasks, each with 5 videos (600 videos in total). For each task, we convert the soft dependency matrix generated by either GPT-4.1-mini or LLaMA-3.3-70B-Instruct into binary (hard) dependencies at various thresholds. To evaluate robustness, we measure the ratio of violated dependencies. A dependency $\dependencymatrix_{i,j}$, indicating that step $\step_j$ is a prerequisite for step $\step_i$, is considered violated if $\step_i$ occurs before the first occurrence of $\step_j$ in any of the videos for that task (\ie, the prerequisite should always occur before the dependent step whenever both are present).

Tab.~\ref{tab:htstep_dependency_violations} reports the percentage of violated dependencies and the percentage of tasks with at least one violation for both LLaMA-3.3-70B-Instruct and GPT-4.1-mini. Both models perform well across different tasks, with roughly 10\% of dependencies being violated depending on the threshold. In particular, LLaMA-3.3-70B-Instruct shows a flat violation ratio ($\sim$8\%) across a wide range of thresholds, suggesting that its outputs are mostly binary (\ie, close to 0 or 1). In contrast, GPT-4.1-mini shows a smoother decline in violation rate as the threshold increases, indicating a more graded confidence distribution. Even at low thresholds, a portion of tasks (28\% for LLaMA-3.3 and 26\% for GPT-4.1-mini at threshold 0.1) have no violated dependencies, suggesting that the predicted matrices can approximate the oracle matrix.

\input{tables/htstep_step_position}
\noindent\textbf{Step prediction by temporal position.} We analyze how step prediction performance varies across the temporal positions of steps in HT-Step. For each step, we normalize its occurrence within the video to the range $[0,1]$ and divide these positions into five bins, then compute R@1 for each bin. 
Tab.~\ref{tab:htstep_step_position} reports R@1 across these bins, from early (bin 0) to late (bin 4) steps. The off-the-shelf model (InternVL2.5-8B), which does not leverage past context, maintains relatively stable performance. It performs better on early and mid-range steps (bin 0: 53.6\%, bin 2: 54.6\%) than on the final steps (bin 4: 49.3\%), suggesting that initial to mid-range steps are more semantically distinctive while later steps are harder to recognize. 

In contrast, \methodshort shows an upward trend from early to mid-range bins, likely due to its use of accumulated temporal context: R@1 increases by +1.1\% from bin 0 to bin 1 and +3.0\% from bin 1 to bin 2. Its performance then decreases toward the final bin, likely due to limitations inherited from the base model (\ie, InternVL2.5-8B performs worst in the final part of the video) as well as error propagation.

\begin{figure}[t]
    \centering
        \includegraphics[width=0.65\linewidth]{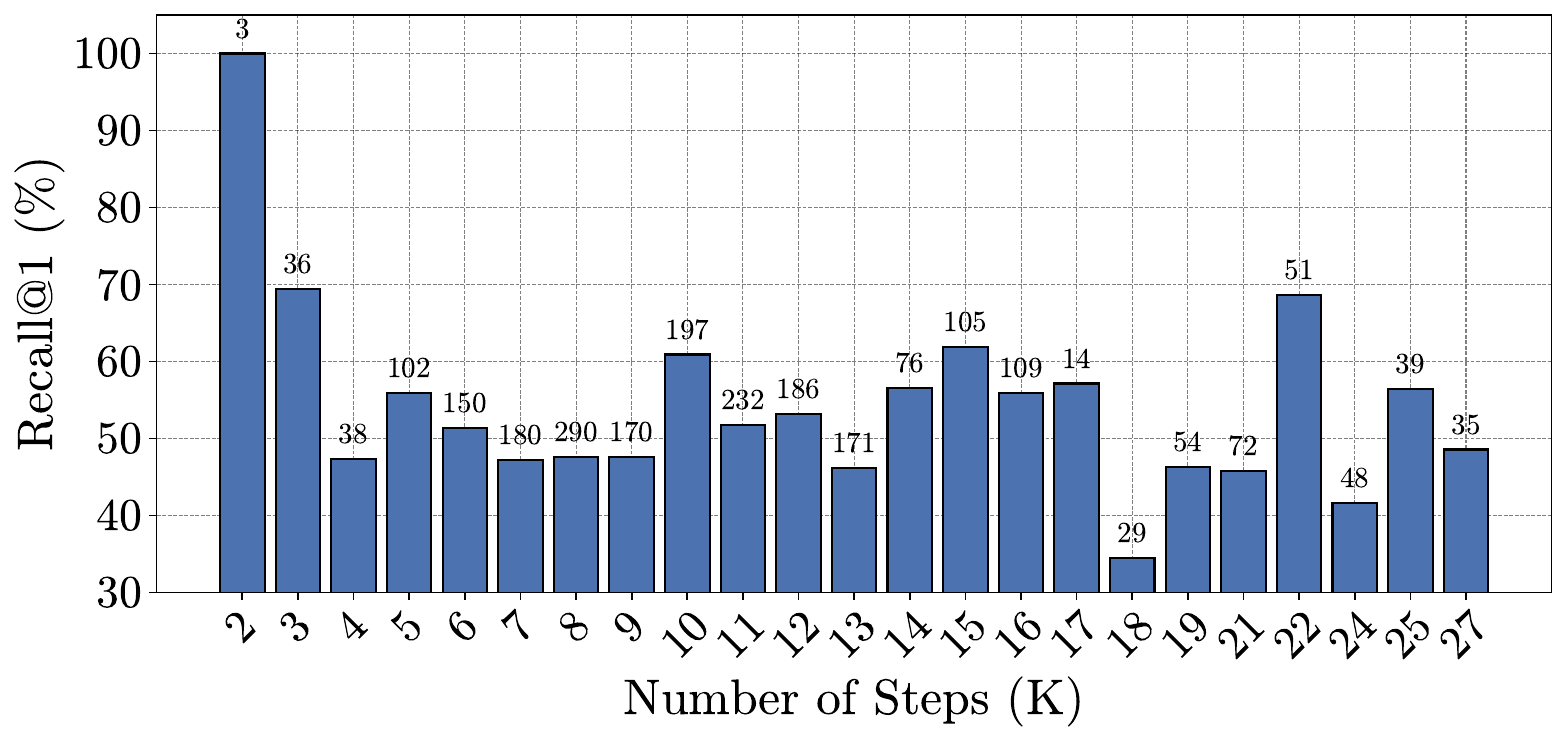}
    \caption{R@1 on the HT-Step dataset, grouped by the number of candidate steps in each task.}
    \label{fig:htstep_recall_by_step_count}
\end{figure}
\noindent\textbf{Step prediction by number of candidate steps.} We analyze how performance changes with the number of candidate steps in the HT-Step dataset by grouping videos based on the number of steps in their tasks. For each group, we compute the average R@1 using the off-the-shelf InternVL2.5-8B model (Fig.~\ref{fig:htstep_recall_by_step_count}). The results show a non-monotonic trend: except for tasks with very few steps (fewer than 4), there is no clear decline in performance as the number of steps increases. This suggests that InternVL2.5-8B can handle prompts with many candidate steps. Note that some bins contain few videos (\eg, 14 and 29 for tasks with 17-18 steps), which can cause large fluctuations in the measurements (\eg, from 57.1\% to 34.5\%).

%% file: tables/ablation_prompt.tex
\begin{table}[t]
\centering
\caption{Comparison of the off-the-shelf InternVL2.5-8B model using different prompts for \taskfull. Queries / $\segment$ indicates the number of LMM queries per video segment $\segment$.}
\label{tab:prompt}
\resizebox{0.7\textwidth}{!}{%
\begin{tabular}{@{}lcccc@{}}
\toprule
\textbf{Prompt} & \textbf{Queries / $\segment$} & \textbf{HT-Step} & \textbf{CrossTask} & \textbf{Ego4D Goal-Step} \\
                &                              & R@1 $\uparrow$   & Avg. R@1 $\uparrow$ & R@1 $\uparrow$          \\
\midrule
\rowcolor{gray!10}
$\pi_{\texttt{VSG}}^\texttt{bin}$ & $\numsteps$ & \textbf{52.5} & \textbf{56.0} & 41.8 \\
$\pi_{\texttt{VSG}}$              & $1$         & 52.0          & 54.3          & \textbf{45.2} \\
\bottomrule
\end{tabular}%
}
\end{table}

%% file: tables/ablation_prior.tex
\begin{table}[t]
\centering
\caption{Ablation study on the step prior $P(\stepvar_t = \step_i)$.}
\label{tab:prior}
\resizebox{0.9\textwidth}{!}{%
\begin{tabular}{@{}lccc@{}}
\toprule
& \textbf{HT-Step} & \textbf{CrossTask} & \textbf{Ego4D Goal-Step} \\
& R@1 $\uparrow$ & Avg. R@1 $\uparrow$ & R@1 $\uparrow$ \\
\midrule
\rowcolor{gray!10} 
1) $f_\mathtt{LMM}(\segment_t, \pi_{\mathtt{next}})[i]$ & 50.5 & 52.6 & 28.1 \\
2) $\sum_{\step_j\in \steps} P(\stepvar_t = \step_i | \stepvar_{t-1} = \step_j)\cdot P(\stepvar_{t-1} = \step_j)$ & 55.6 & 57.3 & 39.9 \\
\rowcolor{gray!10} 
3) $\frac{1}{S+1}$ & \textbf{57.4} & \textbf{59.8} & \textbf{43.3} \\
\bottomrule
\end{tabular}%
}
\end{table}

%% file: tables/htstep_localization.tex
\begin{table}[t]
\centering
\caption{Comparison between \inlineColorbox{gray!10}{NaSVA} and our \inlineColorbox{DrawioBlue}{\methodshort} on the HT-Step dataset.}
\label{tab:htstep_localization}
\resizebox{0.5\textwidth}{!}{%
\begin{tabular}{@{}lcccc@{}}
\toprule
\multirow{2}{*}{\textbf{Method}} & \multicolumn{4}{c}{\textbf{HT-Step} ($\uparrow$ mAP@IoU)} \\
\cmidrule(lr){2-5}
 & @0.3 & @0.5 & @0.7 & @[0.3--0.7] \\
\midrule
\rowcolor{gray!10} \textsc{NaSVA} \citep{li2024multi} & 13.7 & 6.0 & 1.4 & 6.7 \\
\midrule
\rowcolor{DrawioBlue} \textsc{\methodshort} & \textbf{15.9} & \textbf{6.7} & \textbf{1.5} & \textbf{7.7} \\
\bottomrule
\end{tabular}%
}
\end{table}

%% file: tables/ablation_lmm_localization.tex
\begin{table}[t]
\centering
\caption{Ablation study on varying the used LMM on HT-Step. Bold indicates the best result within each LMM group (\ie, with and without \methodshort).}
\label{tab:off_the_shelf_vs_baglm_localization}
\resizebox{0.55\textwidth}{!}{%
\begin{tabular}{@{}lcccc@{}}
\toprule
\multirow{2}{*}{\textbf{Method}} & \multicolumn{4}{c}{\textbf{HT-Step} ($\uparrow$ mAP@IoU)} \\
\cmidrule(lr){2-5}
 & @0.3 & @0.5 & @0.7 & @[0.3--0.7] \\
\midrule
\rowcolor{gray!10} \textsc{LLaVA-OV-7b} \citep{li2025llavaonevision} & 14.1          & 5.5          & 1.3          & 6.6          \\
\rowcolor{gray!10} \hspace{1em}\textbf{+ \methodshort}               & \textbf{14.7} & \textbf{5.8} & \textbf{1.5} & \textbf{6.9} \\
\textsc{Qwen2.5-VL-7b} \citep{bai2025qwen2}                          & 15.5          & 6.1          & 1.4          & 7.3          \\
\hspace{1em}\textbf{+ \methodshort}                                  & \textbf{15.8} & \textbf{6.6} & \textbf{1.7} & \textbf{7.7} \\
\rowcolor{gray!10} \textsc{InternVL3-8b} \citep{zhu2025internvl3}    & 15.6          & 5.8          & 1.4          & 7.2          \\
\rowcolor{gray!10} \hspace{1em}\textbf{+ \methodshort}               & \textbf{17.5} & \textbf{6.5} & \textbf{1.6} & \textbf{8.1} \\
\textsc{InternVL2.5-8b} \citep{chen2024expanding}                    & 15.0          & 6.0          & 1.3          & 7.0          \\
\hspace{1em}\textbf{+ \methodshort}                                  & \textbf{15.9} & \textbf{6.7} & \textbf{1.5} & \textbf{7.7} \\ \bottomrule
\end{tabular}%
}
\end{table}

%% file: tables/coin.tex
\begin{table}[t]
\centering
\caption{Comparison between NaSVA (best \inlineColorbox{gray!10}{offline} competitor) and our \inlineColorbox{DrawioBlue}{online} method \methodshort on the COIN dataset.}
\label{tab:coin}
\resizebox{0.3\textwidth}{!}{%
\begin{tabular}{@{}lc@{}}
\toprule
\textbf{Method}                                                        & \textbf{$\uparrow$ R@1} \\ \midrule
\rowcolor{gray!10}\textit{Offline}                                     &                              \\
\rowcolor{gray!10}\textsc{NaSVA} \citep{li2024multi}                   & 36.8                         \\ \midrule
\rowcolor{DrawioBlue}\textit{Online}                                   &                              \\
\rowcolor{DrawioBlue}\textsc{NaSVA} \citep{li2024multi}                & 32.1                         \\
\rowcolor{DrawioBlue}\textsc{InternVL2.5-8b} \citep{chen2024expanding} & 47.9                         \\
\rowcolor{DrawioBlue}\textbf{\methodshort}                             & \textbf{51.0}                \\ \bottomrule
\end{tabular}%
}
\end{table}

%% file: tables/htstep_prereq_robustness.tex
\begin{table}[t]
\centering
\caption{Dependency violations and fraction of tasks with violations for different threshold values.}
\label{tab:htstep_dependency_violations}
\resizebox{\textwidth}{!}{
\begin{tabular}{l l c c c c c c c c c c c}
\toprule
\textbf{Metric} & \textbf{Model} & 0.00 & 0.10 & 0.20 & 0.30 & 0.40 & 0.50 & 0.60 & 0.70 & 0.80 & 0.90 & 1.00 \\
\midrule
\multirow{2}{*}{Violated dependencies (\%)} 
 & \cellcolor{gray!10}LLaMA-3.3 
   & \cellcolor{gray!10}26.1 
   & \cellcolor{gray!10}8.6 
   & \cellcolor{gray!10}8.4 
   & \cellcolor{gray!10}8.4 
   & \cellcolor{gray!10}8.4 
   & \cellcolor{gray!10}8.4 
   & \cellcolor{gray!10}8.4 
   & \cellcolor{gray!10}8.3 
   & \cellcolor{gray!10}8.3 
   & \cellcolor{gray!10}8.1 
   & \cellcolor{gray!10}8.1 \\
 & GPT-4.1-mini 
   & 26.1 & 15.4 & 12.5 & 11.6 & 10.8 & 10.2 & 9.6 & 8.9 & 8.2 & 7.6 & 0.0 \\
\midrule
\multirow{2}{*}{Tasks w/ violation (\%)} 
 & \cellcolor{gray!10}LLaMA-3.3 
   & \cellcolor{gray!10}100.0 
   & \cellcolor{gray!10}72.0 
   & \cellcolor{gray!10}71.3 
   & \cellcolor{gray!10}71.3 
   & \cellcolor{gray!10}69.3 
   & \cellcolor{gray!10}69.3 
   & \cellcolor{gray!10}69.3 
   & \cellcolor{gray!10}68.7 
   & \cellcolor{gray!10}68.7 
   & \cellcolor{gray!10}68.7 
   & \cellcolor{gray!10}68.7 \\
 & GPT-4.1-mini 
   & 100.0 & 74.0 & 65.3 & 64.0 & 59.3 & 54.0 & 51.3 & 48.0 & 44.7 & 42.0 & 0.0 \\
\bottomrule
\end{tabular}}
\end{table}

%% file: tables/htstep_step_position.tex
\begin{table}[t]
\centering
\caption{R@1 performance across step positions on HT-Step.}
\label{tab:htstep_step_position}
\resizebox{0.5\textwidth}{!}{%
\begin{tabular}{@{}lccccc@{}}
\toprule
\multicolumn{1}{c}{\textbf{Model}} & \textbf{Bin 0} & \textbf{Bin 1} & \multicolumn{1}{l}{\textbf{Bin 2}} & \multicolumn{1}{l}{\textbf{Bin 3}} & \textbf{Bin 4} \\ \midrule
InternVL2.5-8b                     & 53.6           & 52.7           & 54.6                               & 49.7                               & 49.3           \\
\rowcolor{DrawioBlue}\methodshort  & 56.1           & 57.2           & 60.2                               & 56.8                               & 56.6           \\ \bottomrule
\end{tabular}%
}
\end{table}

%% file: appendix/c_inference.tex
\section{Inference speed analysis}
\label{sec:app:inference}

\begin{figure}[t]
    \centering
    \includegraphics[width=0.65\linewidth]{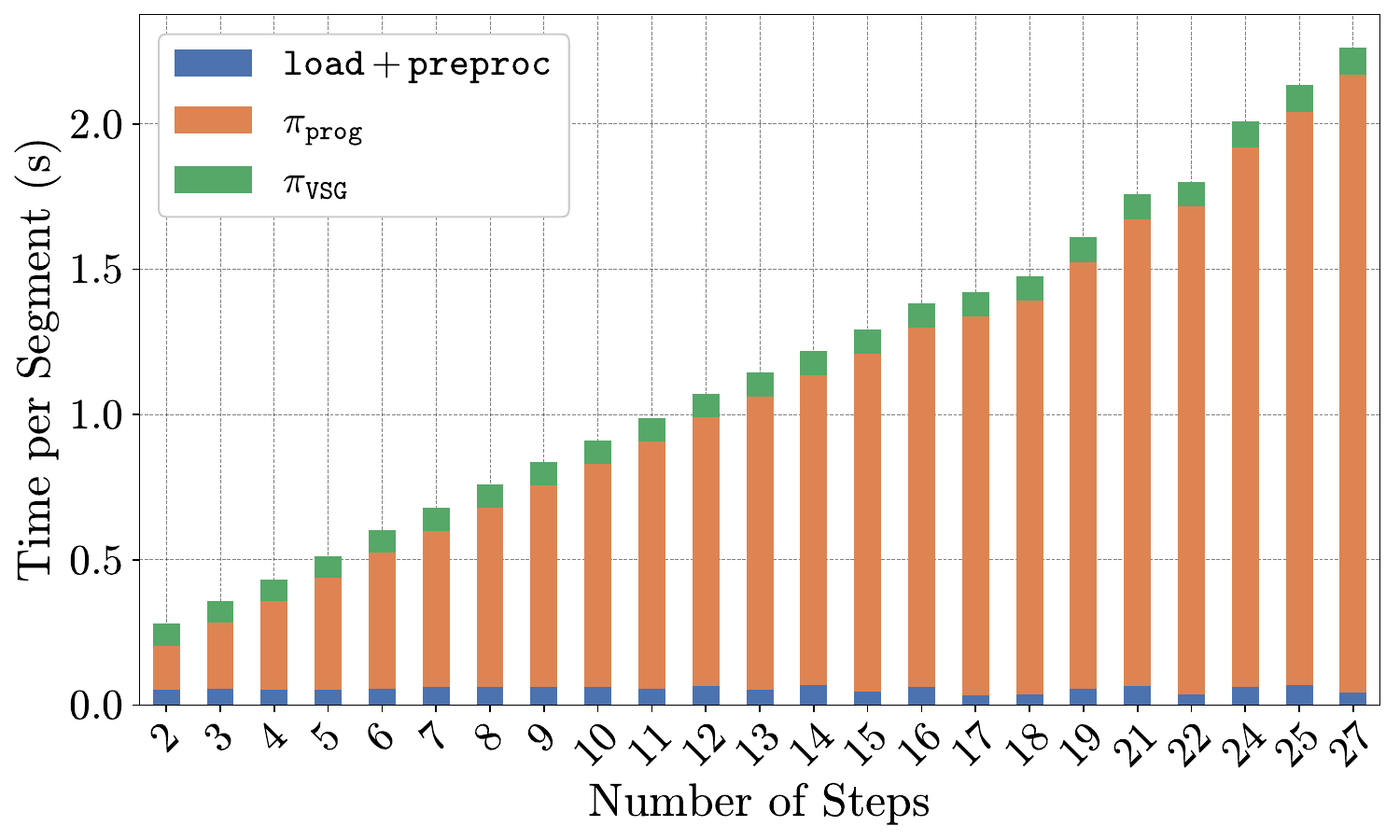}
    \caption{Average processing time per video segment for computing \taskshort scores, estimating action progress, and video loading/preprocessing, for HT-Step videos grouped by their number of steps.}
    \label{fig:inference_stats}
\end{figure}

We measure inference speed on a single NVIDIA H100 GPU with 64GB of memory.  
Figure~\ref{fig:inference_stats} shows the average processing time per video segment for three main operations: video loading and preprocessing, computing \taskshort scores, and estimating action progress.
These estimates were computed on HT-Step videos with the same number of steps.  
Our results show that the total processing time per segment is largely dominated by action progress estimation, which grows linearly with the number of steps in the task. Importantly, the majority of the segments are processed in under 2 seconds, implying that the 2-second segments can be processed in real time.

As an example, consider the video \texttt{-25-1nbOki0} from the HT-Step dataset.  This video is 567 seconds long and corresponds to the task \textit{Make Nigerian Style Jollof Rice}, which includes 10 steps. We generate the step dependency matrix for this task using GPT-4.1-mini with prompt $\pi_{\mathtt{prereq}}$, requiring 87.84 seconds to compute.  This matrix is generated only once per task and reused for all related videos.  We process each 2-second video segment as follows: we compute \taskshort scores using InternVL2.5-8B with prompt $\pi_{\texttt{VSG}}$, taking 0.088 seconds per segment. We estimate action progress using prompt $\pi_{\mathtt{prog}}$, requiring 0.814 seconds per segment. Additional time is spent on model initialization (performed once) and video decoding and preprocessing. Using the \texttt{torchcodec}\footnote{\url{https://docs.pytorch.org/torchcodec/stable/index.html}} library, decoding and preprocessing take approximately 1.201 seconds per segment on CPU, with possible speedups using GPU acceleration. In comparison, the Bayesian filtering step is computationally negligible. Overall, excluding step dependency generation and model loading, each 2-second segment of this video can be processed in approximately 2.1 seconds.

%% file: appendix/d_broader_impacts.tex
\section{Broader impact}
\label{sec:app:broader_impacts}

Our approach enables online understanding of instructional video steps without requiring task-specific training. This makes it suitable for deployment in real environments, providing feedback to users performing tasks. Such capabilities can benefit a wide range of applications, including education, skill development, and assistive technologies, especially in contexts where annotated data is limited or users require immediate feedback (\eg, remote learning, home cooking, or physical rehabilitation). However, as our method depends on large pretrained multimodal models, it may also inherit their limitations, such as biased predictions or inconsistent performance across different cultural or domain-specific tasks. Future work should explore strategies for mitigating such biases, ensuring equal performance across diverse user populations.

%% file: appendix/e_assets.tex
\section{Assets}
\label{sec:app:assets}

\input{tables/assets}

Tab.~\ref{tab:assets} lists URLs and licenses for all the assets used in the paper. 

%% file: tables/assets.tex
\begin{table}[t]
\centering
\caption{List of URLs and licenses for all assets used.}
\label{tab:assets}
\resizebox{\textwidth}{!}{%
\begin{tabular}{@{}lll@{}}
\toprule
\textbf{Name} & \textbf{URL} & \textbf{License} \\ \midrule
\textsc{LLaVA-OV-7B} & \url{https://huggingface.co/lmms-lab/llava-onevision-qwen2-7b-ov} & Apache 2.0 \\
\textsc{Qwen2.5-VL-7B} & \url{https://huggingface.co/Qwen/Qwen2.5-VL-7B-Instruct} & Apache 2.0 \\
\textsc{InternVL2.5-8B} & \url{https://huggingface.co/OpenGVLab/InternVL2_5-8B} & MIT \\
\textsc{InternVL3-8B} & \url{https://huggingface.co/OpenGVLab/InternVL3-8B} & MIT \\
\textsc{CrossTask} & \url{https://github.com/DmZhukov/CrossTask} & BSD-3-Clause \\
\textsc{HT-Step} & \url{https://github.com/facebookresearch/htstep} & CC-BY-NC 4.0 \\
\textsc{Ego4D Goal-Step} & \url{https://github.com/facebookresearch/ego4d-goalstep} & MIT \\
\textsc{COIN} & \url{https://github.com/coin-dataset/annotations} & CC BY-NC 4.0 \\
\bottomrule
\end{tabular}%
}
\end{table}

%% file: appendix/f_qualitatives.tex
\section{Qualitative results}
\label{sec:app:qualitatives}

\begin{figure}[t]
    \centering
    \includegraphics[width=0.82\linewidth]{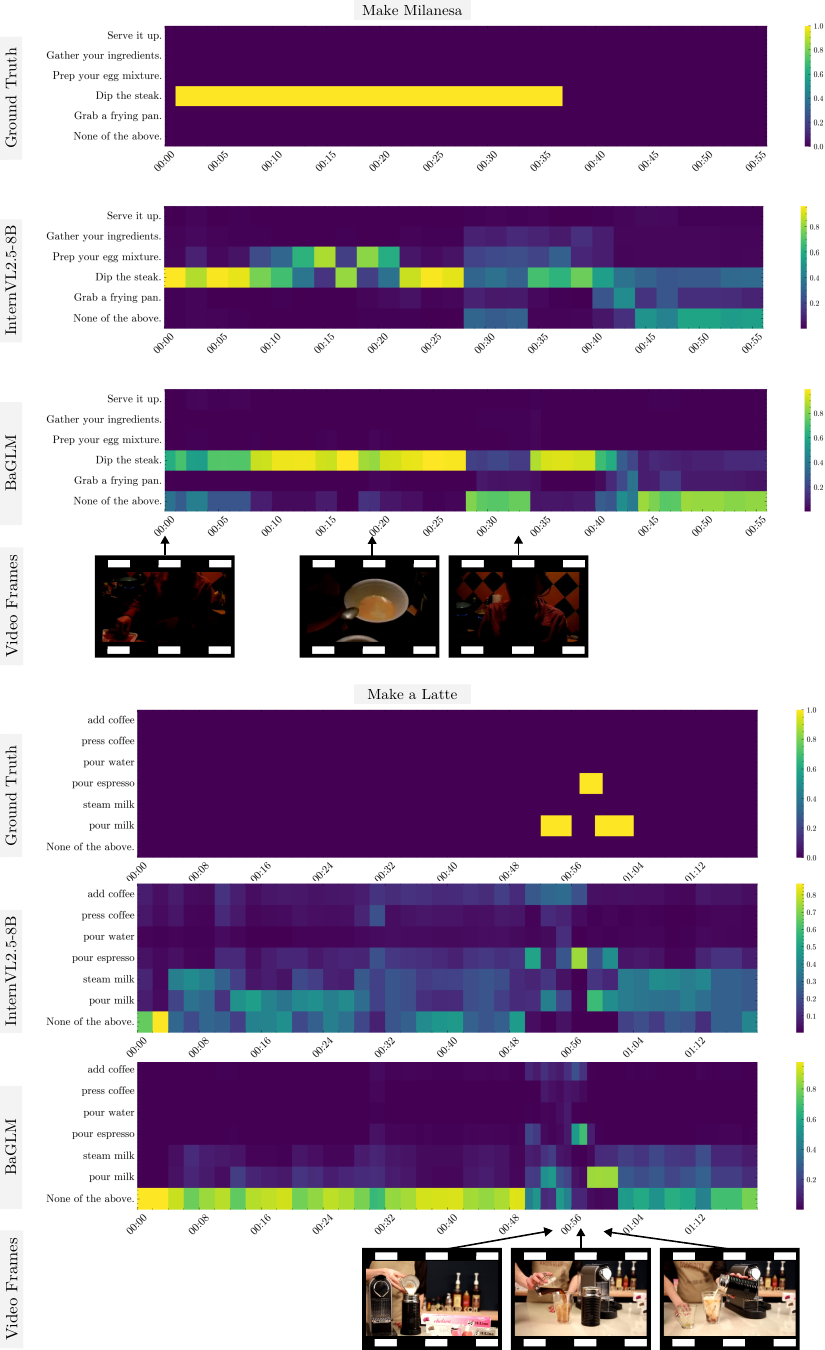}
    \caption{Qualitative results of \methodshort on test videos from HT-Step (\textit{Make Milanesa}) and CrossTask (\textit{Make a Latte}). Ground truth step boundaries and predicted step probabilities per segment are shown for both \methodshort and the off-the-shelf LMM. Arrows point to the timestamps of selected keyframes.}
    \label{fig:qualitatives}
\end{figure}

Fig.~\ref{fig:qualitatives} showcases qualitative results produced by \methodshort on two test videos: one from the HT-Step dataset (task: \textit{Make Milanesa}) and one from the CrossTask dataset (task: \textit{Make a Latte}). For each video, we plot the ground truth step boundaries and the step probabilities over video segments, as predicted by our method and by the off-the-shelf version of the LMM. We also show keyframes at selected time points, indicated by arrows.

In the \textit{Make Milanesa} video, the off-the-shelf LMM assigns a high probability to the step \textit{Dip the steak} at the very beginning of the video, as the person is seen picking up the steak from a plate, as shown by the corresponding keyframe. For the same step, \methodshort initially assigns a low probability but increases it over time, consistent with the true presence of the step in the video. 
Around timestamp \texttt{00:20}, the off-the-shelf LMM mistakenly assigns a high probability to step \textit{Prep your egg mixture}, likely due to the presence of the egg mixture in the video segment. In contrast, \methodshort consistently predicts \textit{Dip the steak} at this point, leveraging prior context to avoid a false positive. At timestamp \texttt{00:30}, although the ground truth still indicates \textit{Dip the steak} is in progress, the step is already completed and \methodshort correctly detects this.

In the \textit{Make a Latte} video, \methodshort effectively suppresses probabilities for steps that fall outside their actual temporal boundaries. Leveraging LMMs to compute \taskshort scores proves particularly beneficial in distinguishing visually similar actions such as \textit{pour milk} and \textit{pour espresso}. Importantly, steps not present in the video are not falsely identified by \methodshort, showcasing its precision.

%% file: neurips_2025.bbl
\begin{thebibliography}{10}

\bibitem{afouras2023ht}
Triantafyllos Afouras, Effrosyni Mavroudi, Tushar Nagarajan, Huiyu Wang, and Lorenzo Torresani.
\newblock Ht-step: Aligning instructional articles with how-to videos.
\newblock {\em NeurIPS}, 2023.

\bibitem{bai2025qwen2}
Shuai Bai, Keqin Chen, Xuejing Liu, Jialin Wang, Wenbin Ge, Sibo Song, Kai Dang, Peng Wang, Shijie Wang, Jun Tang, et~al.
\newblock Qwen2. 5-vl technical report.
\newblock {\em arXiv}, 2025.

\bibitem{chen2021multimodal}
Brian Chen, Andrew Rouditchenko, Kevin Duarte, Hilde Kuehne, Samuel Thomas, Angie Boggust, Rameswar Panda, Brian Kingsbury, Rogerio Feris, David Harwath, et~al.
\newblock Multimodal clustering networks for self-supervised learning from unlabeled videos.
\newblock In {\em ICCV}, 2021.

\bibitem{chen2024learning}
Yuxiao Chen, Kai Li, Wentao Bao, Deep Patel, Yu~Kong, Martin~Renqiang Min, and Dimitris~N Metaxas.
\newblock Learning to localize actions in instructional videos with llm-based multi-pathway text-video alignment.
\newblock In {\em ECCV}, 2024.

\bibitem{chen2003bayesian}
Zhe Chen.
\newblock Bayesian filtering: From kalman filters to particle filters, and beyond.
\newblock {\em Statistics}, 2003.

\bibitem{chen2024expanding}
Zhe Chen, Weiyun Wang, Yue Cao, Yangzhou Liu, Zhangwei Gao, Erfei Cui, Jinguo Zhu, Shenglong Ye, Hao Tian, Zhaoyang Liu, et~al.
\newblock Expanding performance boundaries of open-source multimodal models with model, data, and test-time scaling.
\newblock {\em arXiv}, 2024.

\bibitem{grattafiori2024llama}
Aaron Grattafiori, Abhimanyu Dubey, Abhinav Jauhri, Abhinav Pandey, Abhishek Kadian, Ahmad Al-Dahle, Aiesha Letman, Akhil Mathur, Alan Schelten, Alex Vaughan, et~al.
\newblock The llama 3 herd of models.
\newblock {\em arXiv}, 2024.

\bibitem{grauman2022ego4d}
Kristen Grauman, Andrew Westbury, Eugene Byrne, Zachary Chavis, Antonino Furnari, Rohit Girdhar, Jackson Hamburger, Hao Jiang, Miao Liu, Xingyu Liu, et~al.
\newblock Ego4d: Around the world in 3,000 hours of egocentric video.
\newblock In {\em CVPR}, 2022.

\bibitem{han2022temporal}
Tengda Han, Weidi Xie, and Andrew Zisserman.
\newblock Temporal alignment networks for long-term video.
\newblock In {\em CVPR}, 2022.

\bibitem{hessel2021clipscore}
Jack Hessel, Ari Holtzman, Maxwell Forbes, Ronan~Le Bras, and Yejin Choi.
\newblock Clipscore: A reference-free evaluation metric for image captioning.
\newblock {\em arXiv}, 2021.

\bibitem{ko2022video}
Dohwan Ko, Joonmyung Choi, Juyeon Ko, Shinyeong Noh, Kyoung-Woon On, Eun-Sol Kim, and Hyunwoo~J Kim.
\newblock Video-text representation learning via differentiable weak temporal alignment.
\newblock In {\em CVPR}, 2022.

\bibitem{li2024evaluating}
Baiqi Li, Zhiqiu Lin, Deepak Pathak, Jiayao Li, Yixin Fei, Kewen Wu, Xide Xia, Pengchuan Zhang, Graham Neubig, and Deva Ramanan.
\newblock Evaluating and improving compositional text-to-visual generation.
\newblock In {\em CVPR}, 2024.

\bibitem{li2025llavaonevision}
Bo~Li, Yuanhan Zhang, Dong Guo, Renrui Zhang, Feng Li, Hao Zhang, Kaichen Zhang, Peiyuan Zhang, Yanwei Li, Ziwei Liu, and Chunyuan Li.
\newblock {LL}a{VA}-onevision: Easy visual task transfer.
\newblock {\em Transactions on Machine Learning Research}, 2025.

\bibitem{li2024multi}
Zeqian Li, Qirui Chen, Tengda Han, Ya~Zhang, Yanfeng Wang, and Weidi Xie.
\newblock Multi-sentence grounding for long-term instructional video.
\newblock In {\em ECCV}, 2024.

\bibitem{lin2023revisiting}
Zhiqiu Lin, Xinyue Chen, Deepak Pathak, Pengchuan Zhang, and Deva Ramanan.
\newblock Revisiting the role of language priors in vision-language models.
\newblock {\em arXiv}, 2023.

\bibitem{lin2024evaluating}
Zhiqiu Lin, Deepak Pathak, Baiqi Li, Jiayao Li, Xide Xia, Graham Neubig, Pengchuan Zhang, and Deva Ramanan.
\newblock Evaluating text-to-visual generation with image-to-text generation.
\newblock In {\em ECCV}. Springer, 2024.

\bibitem{luo2020univl}
Huaishao Luo, Lei Ji, Botian Shi, Haoyang Huang, Nan Duan, Tianrui Li, Jason Li, Taroon Bharti, and Ming Zhou.
\newblock Univl: A unified video and language pre-training model for multimodal understanding and generation.
\newblock {\em arXiv}, 2020.

\bibitem{mavroudi2023learning}
Effrosyni Mavroudi, Triantafyllos Afouras, and Lorenzo Torresani.
\newblock Learning to ground instructional articles in videos through narrations.
\newblock In {\em ICCV}, 2023.

\bibitem{miech2020end}
Antoine Miech, Jean-Baptiste Alayrac, Lucas Smaira, Ivan Laptev, Josef Sivic, and Andrew Zisserman.
\newblock End-to-end learning of visual representations from uncurated instructional videos.
\newblock In {\em CVPR}, 2020.

\bibitem{miech2019howto100m}
Antoine Miech, Dimitri Zhukov, Jean-Baptiste Alayrac, Makarand Tapaswi, Ivan Laptev, and Josef Sivic.
\newblock Howto100m: Learning a text-video embedding by watching hundred million narrated video clips.
\newblock In {\em ICCV}, 2019.

\bibitem{gpt4.1-mini}
OpenAI.
\newblock Gpt-4o mini: advancing cost-efficient intelligence.
\newblock \url{https://openai.com/index/gpt-4-1/}, 2024.

\bibitem{plizzari2025omnia}
Chiara Plizzari, Alessio Tonioni, Yongqin Xian, Achin Kulshrestha, and Federico Tombari.
\newblock Omnia de egotempo: Benchmarking temporal understanding of multi-modal llms in egocentric videos.
\newblock {\em CVPR}, 2025.

\bibitem{radford2021learning}
Alec Radford, Jong~Wook Kim, Chris Hallacy, Aditya Ramesh, Gabriel Goh, Sandhini Agarwal, Girish Sastry, Amanda Askell, Pamela Mishkin, Jack Clark, et~al.
\newblock Learning transferable visual models from natural language supervision.
\newblock In {\em ICML}, 2021.

\bibitem{shen2024progress}
Yuhan Shen and Ehsan Elhamifar.
\newblock Progress-aware online action segmentation for egocentric procedural task videos.
\newblock In {\em CVPR}, 2024.

\bibitem{shen2021learning}
Yuhan Shen, Lu~Wang, and Ehsan Elhamifar.
\newblock Learning to segment actions from visual and language instructions via differentiable weak sequence alignment.
\newblock In {\em CVPR}, 2021.

\bibitem{shi2022emscore}
Yaya Shi, Xu~Yang, Haiyang Xu, Chunfeng Yuan, Bing Li, Weiming Hu, and Zheng-Jun Zha.
\newblock Emscore: Evaluating video captioning via coarse-grained and fine-grained embedding matching.
\newblock In {\em CVPR}, 2022.

\bibitem{song2023ego4d}
Yale Song, Eugene Byrne, Tushar Nagarajan, Huiyu Wang, Miguel Martin, and Lorenzo Torresani.
\newblock Ego4d goal-step: Toward hierarchical understanding of procedural activities.
\newblock {\em NeurIPS}, 2023.

\bibitem{tang2019coin}
Yansong Tang, Dajun Ding, Yongming Rao, Yu~Zheng, Danyang Zhang, Lili Zhao, Jiwen Lu, and Jie Zhou.
\newblock Coin: A large-scale dataset for comprehensive instructional video analysis.
\newblock In {\em CVPR}, 2019.

\bibitem{wikihow}
wikiHow.
\newblock wikihow: How-to instructions you can trust.
\newblock \url{https://www.wikihow.com/}.

\bibitem{wu2024towards}
Jay~Zhangjie Wu, Guian Fang, Haoning Wu, Xintao Wang, Yixiao Ge, Xiaodong Cun, David~Junhao Zhang, Jia-Wei Liu, Yuchao Gu, Rui Zhao, et~al.
\newblock Towards a better metric for text-to-video generation.
\newblock {\em arXiv}, 2024.

\bibitem{wu2024vila}
Yecheng Wu, Zhuoyang Zhang, Junyu Chen, Haotian Tang, Dacheng Li, Yunhao Fang, Ligeng Zhu, Enze Xie, Hongxu Yin, Li~Yi, et~al.
\newblock Vila-u: a unified foundation model integrating visual understanding and generation.
\newblock {\em arXiv}, 2024.

\bibitem{xu2021videoclip}
Hu~Xu, Gargi Ghosh, Po-Yao Huang, Dmytro Okhonko, Armen Aghajanyan, Florian Metze, Luke Zettlemoyer, and Christoph Feichtenhofer.
\newblock Videoclip: Contrastive pre-training for zero-shot video-text understanding.
\newblock In {\em EMNLP}, 2021.

\bibitem{zanella2025can}
Luca Zanella, Massimiliano Mancini, Willi Menapace, Sergey Tulyakov, Yiming Wang, and Elisa Ricci.
\newblock Can text-to-video generation help video-language alignment?
\newblock {\em CVPR}, 2025.

\bibitem{zhang2021natural}
Hao Zhang, Aixin Sun, Wei Jing, Liangli Zhen, Joey~Tianyi Zhou, and Rick Siow~Mong Goh.
\newblock Natural language video localization: A revisit in span-based question answering framework.
\newblock {\em IEEE TPAMI}, 2021.

\bibitem{zhu2025internvl3}
Jinguo Zhu, Weiyun Wang, Zhe Chen, Zhaoyang Liu, Shenglong Ye, Lixin Gu, Yuchen Duan, Hao Tian, Weijie Su, Jie Shao, et~al.
\newblock Internvl3: Exploring advanced training and test-time recipes for open-source multimodal models.
\newblock {\em arXiv}, 2025.

\bibitem{zhukov2019cross}
Dimitri Zhukov, Jean-Baptiste Alayrac, Ramazan~Gokberk Cinbis, David Fouhey, Ivan Laptev, and Josef Sivic.
\newblock Cross-task weakly supervised learning from instructional videos.
\newblock In {\em CVPR}, 2019.

\end{thebibliography}
